\pdfoutput=1

\documentclass[11pt]{article}

\usepackage[preprint]{acl}

\usepackage{times}
\usepackage{latexsym}

\usepackage[T1]{fontenc}

\usepackage[utf8]{inputenc}

\usepackage{microtype}

\usepackage{inconsolata}

\usepackage{wrapfig}
\usepackage{subcaption}
\usepackage{graphicx}
\usepackage{enumitem}
\usepackage{amsmath}   
\usepackage{amssymb}   
\usepackage{bm}        
\usepackage{booktabs}   
\usepackage{soul} 
\usepackage{multirow} 
\usepackage{pifont}
\usepackage{longtable}
\usepackage{svg}
\svgpath{{../img/}}

\definecolor{LightRed}{HTML}{ffdfd5}
\definecolor{LightYellow}{HTML}{ffef95}
\sethlcolor{LightRed}

\expandafter\def\expandafter\normalsize\expandafter{%
    \normalsize%
    \setlength\abovedisplayskip{3pt}%
    \setlength\belowdisplayskip{3pt}%
    \setlength\abovedisplayshortskip{0pt}%
    \setlength\belowdisplayshortskip{0pt}%
}

\newcommand{\modelname}{PERFT}
\newcommand{\mixtral}{Mixtral-8$\times$7B}
\newcommand{\olmoe}{OLMoE-1B-7B}

\newcommand{\rmv}[1]{}
\newcommand{\cc}{\cellcolor{LightRed}}

\title{Parameter-Efficient Routed Fine-Tuning: 
\\Mixture-of-Experts Demands Mixture of Adaptation Modules}

\author{
\textbf{Yilun Liu}${}^{1,2,\dagger}$ \quad
\textbf{Yunpu Ma}${}^{2,\dagger}$ \quad
\textbf{Yuetian Lu}${}^{1}$ \quad \\
\textbf{Shuo Chen}${}^{2}$ \quad
\textbf{Zifeng Ding}${}^{2,3}$ \quad
\textbf{Volker Tresp}${}^{2,\dagger}$\\
${}^{1}$\text{Technical University of Munich} \quad
${}^{2}$\text{Ludwig Maximilian University of Munich} \\
${}^{3}$\text{University of Cambridge} \\
\normalsize
${}^{\dagger}$\texttt{yilun.liu@tum.de} 
\quad \texttt{cognitive.yunpu@gmail.com}
\quad \texttt{volker.tresp@lmu.de}\\}

\begin{document}
\maketitle
\begin{abstract}
Mixture-of-Experts (MoE) benefits from a dynamic routing mechanism among their specialized experts, which existing Parameter-Efficient Fine-Tuning (PEFT) strategies fail to leverage.
This motivates us to investigate whether adaptation modules themselves should incorporate routing mechanisms to align with MoE's multi-expert architecture.
We analyze dynamics of core components when applying PEFT to MoE language models and examine how different routing strategies affect adaptation effectiveness. 
Extensive experiments adapting \olmoe{} and \mixtral{} on various commonsense and math reasoning tasks validate the performance and efficiency of our routed approach. 
We identify the optimal configurations for different scenarios and provide empirical analyses with practical insights to facilitate better PEFT and MoE applications.\footnote{Code available at \url{https://anonymous.4open.science/r/PERFT-MoE/}.}
\end{abstract}

\rmv{
observation
research question
design our approach & exp to validate the RQ
findings

\vspace{-0.1cm}
Mixture-of-Experts (MoE) has become a popular large language model (LLM) architecture. 
However, efficiently fine-tuning MoE LLMs is still an open challenge.
While Parameter-Efficient Fine-Tuning (PEFT) has been widely adopted for dense LLMs, its adaptation to MoE LLMs remains underexplored.
This work investigates the dynamics of core components when PEFT-ing MoE LLMs, and introduces a unified framework for integrating PEFT into MoE LLMs, with a comprehensive set of functional and composition design dimensions to exploit MoE's inherent flexibility.
We further propose a flexible and scalable family of PEFT strategies tailored for MoE LLMs.  
Extensive experiments adapting \olmoe{} and \mixtral{} on various commonsense and math reasoning tasks validate the performance and efficiency of our solutions. 
We identify the optimal configurations for each dimension and provide empirical analyses with practical insights to facilitate better PEFT and MoE applications. \footnote{Code available via \url{https://anonymous.4open.science/r/PERFT-MoE/}.}
}
\begin{figure*}[t]
    \centering
\includegraphics[width=\linewidth]{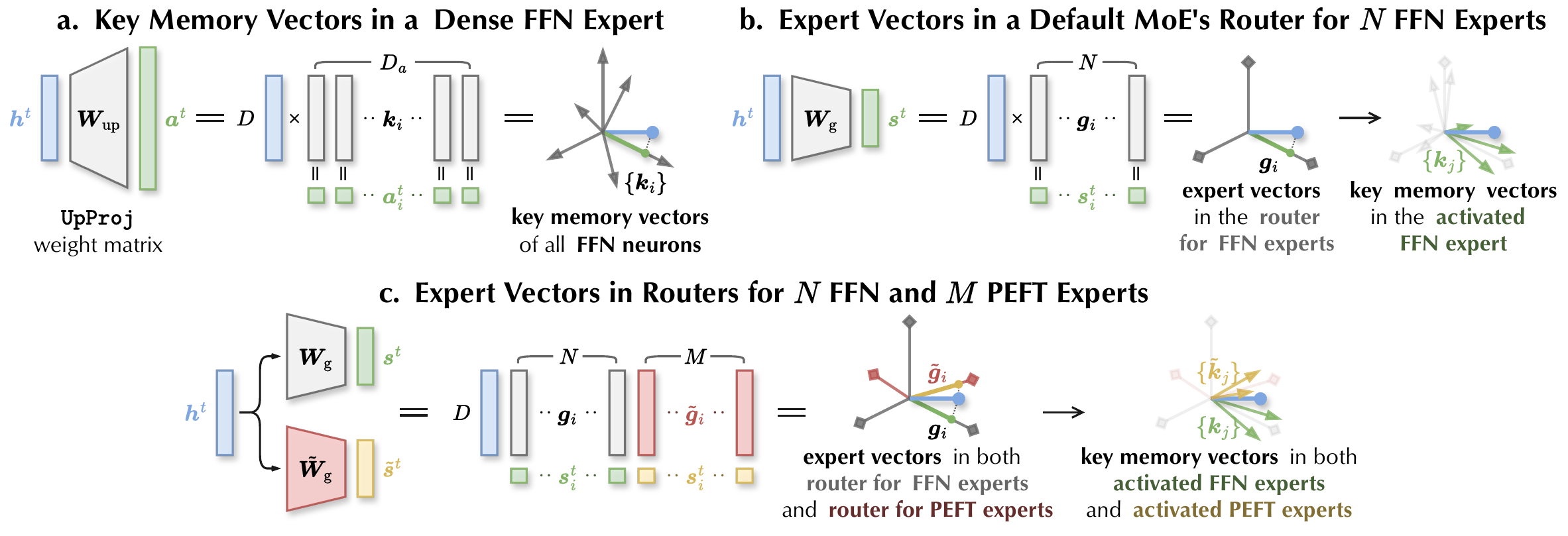}
    \caption{
\textbf{Dynamics between key memory vectors in experts and expert vectors in routers.} 
\textbf{a.}
Dense Feed-Forward Network (FFN) projects hidden state \(\bm{h}^t\in\mathbb{R}^D\) onto \(D_a\) key memory vectors \(\bm{k}_i\in\mathbb{R}^D\) in weight matrix \(\bm{W}_\text{up}\), yielding activation scores \(\bm{a}^t\in\mathbb{R}^{D_a}\). 
\textbf{b.}
Router for $N$ FFN experts projects \(\bm{h}^t\) onto \(N\) expert vectors \(\bm{g}_i\in\mathbb{R}^D\) in router weight matrix \(\bm{W}_g\), yielding token-to-expert affinity scores \(\bm{s}^t\in\mathbb{R}^{N}\).
Each $\bm{g}_i$ symbolizes a characteristic \(\bm{h}^t\) pattern with the activation of corresponding expert's $\bm{k}_j$.
\textbf{c.}
Routers for both FFN and PEFT experts introduce interesting dynamics among their expert vectors, resulting a flexible space for fine-tuning.
}
\label{fig:router}
\end{figure*}

\section{Introduction}

As modern transformer-based large language models (LLMs) continue to scale \citep{vaswani2017attention}, Mixture-of-Experts (MoE) has emerged as a promising approach \citep{shazeer2017outrageously}, powering series of frontier models \citep{jiang2024mixtral, qwen_moe, deepseekR1}. Fine‑tuning these sparse yet massive models poses unique challenges, that direct full fine‑tuning is not only expensive but ignores the routed dynamics and sparsity of experts, negating their computational advantages \citep{wang2024let}. 
Existing Parameter-Efficient Fine-Tuning (PEFT) strategies like LoRA (Low-Rank Adaptation) have been widely studied on dense LLMs \citep{houlsby2019parameter, hu2021lora, he2021towards}. Yet directly adapting MoE LLMs with PEFT is not an ideal solution, since current practice often treats MoE as dense and only addresses MoE-irrelevant modules.

These observations motivate us to investigate the designs for PEFT modules that consider the underlying routing mechanisms of MoE. 
Recent studies have explored \emph{MoE‑inspired} PEFT modules targeting dense backbones \citep{zadouri2023molora,li2024mixlora,hao2024meft}, which inspired us to propose that a mixture of PEFT modules should be similarly required for adapting MoE LLMs.

To verify this, we start by analyzing the dynamics between \emph{key memory vectors} \citep{geva2020transformer} in experts and \emph{expert vectors} in routers.
In §\ref{dynamics} and Figure \ref{fig:router}, We demonstrate that properly routed PEFT experts can unlock a much more expressive adaptation space while maintaining MoE's efficiency and flexibility.

\rmv{and fine‑tunes a subset of pretrained experts without keeping original weights untouched .\citep{wang2024let}}

Guided by these insights, we introduce a framework to explore meaningful design choices for integrating PEFT modules into MoE LLMs in §\ref{framework} and Figure \ref{fig:framework}. We define (i) \emph{functional} strategies, including the architecture, multiplicity, routing among PEFT experts; and (ii) \emph{compositional} strategies, specifying how PEFT modules interact with the original MoE module. 
Within this framework, we further propose \textbf{P}arameter‑\textbf{E}fficient \textbf{R}outed \textbf{F}ine‑\textbf{T}uning (\textbf{\textsc{Perft}}) and three ablated variants (\textsc{Perft}‑E/D/S) in §\ref{strategies} and Figure \ref{fig:architecture}. 
These strategies allow us to systematically verify if MoE actually demands a mixture of adaptation modules.

We evaluate our proposed strategies on \olmoe{} \citep{muennighoff2024olmoe} and \mixtral{} \citep{jiang2024mixtral} across 14 commonsense and arithmetic reasoning tasks. 
\textsc{Perft} yields up to 17.2\% relative improvement over MoE‑agnostic baselines with equivalent number of activated parameters, showing that mixture of adaptation modules can indeed achieve better results on MoE LLMs. 
We also systematically explore the optimal scaling, sparsity, and routing configurations, and empirically analyzed our findings with insights that generalize across settings and may facilitate better PEFT and MoE applications.

The primary contributions of this paper are:
\begin{enumerate}[leftmargin=0.5cm, itemsep=0pt, topsep=2pt]
\item \textbf{Dynamics} between experts and routers when applying PEFT to MoE LLMs;
\item \textbf{Framework \& Strategies} for systematic exploration of PEFT design choices;
\item \textbf{Evidence \& Guidelines} for the gains of routed adaptation strategies on MoE LLMs.
\end{enumerate}

\rmv{
\noindent\textbf{Novelty at a glance:}
\begin{enumerate}[leftmargin=0.75cm, itemsep=0pt, topsep=2pt]
    \item \textbf{Beyond MoE‑LoRA / MixLoRA.} Those methods add LoRA-Mixtures \emph{inside dense} LLMs; we operate on \emph{already sparse} MoE LLMs and attach PEFT experts without destroying their token‑wise sparsity.
    \item \textbf{Complementary to expert‑specialised FT \citep{wang2024let}.} Whereas that work unfreezes a handful of pretrained experts, we \emph{keep all pretrained weights frozen} and add lightweight experts, delivering higher parameter efficiency and eliminating forgetting.
    \item \textbf{Different focus from Self‑MoE \citep{kang2024selfmoe}.} Self‑MoE creates LoRA experts \emph{and trains a new router} on a dense base model\footnote{\url{https://arxiv.org/abs/2406.12034}}; our \textsc{\textsc{Perft}‑R} plugs PEFT experts into an existing MoE LLM and can either \emph{reuse} the pretrained router (\textsc{\textsc{Perft}‑E}) or \emph{learn} an independent router, achieving task‑adaptive sparsity (§\ref{exp}).
\end{enumerate}
}
\section{Methodology}
\label{method}

\rmv{
2.1 shows that Mixture-of adapation integrated with MoE COULD improve performance
2.2 we provide framework to explore how such integration can take place
2.3 meaningful designs are implemented in detail for experimenting
}

We start from investigating how the core components of MoE and PEFT modules interact, which creates new opportunities for designing PEFT on MoE LLMs.
\rmv{
\begin{table}[t]
\centering
\small
\setlength{\tabcolsep}{4pt}
\begin{tabular}{@{}l m{0.45\linewidth}@{}}
\toprule
\textbf{Term} & \textbf{Definition (this paper)} \\ \midrule
Key memory vector $\bm{k}_i$ &
$\,i$‑th column of $\bm W_{\text{up}}$ in an FFN; acts as a pattern detector and produces activation scores.\\
Expert vector $\bm{g}_i$ &
$\,i$‑th column of router weight $\bm W_g$; projects a token $\bm h^t$ to an affinity score for expert~$E_i$.\\
PEFT expert $\Delta_j(\cdot)$ &
A lightweight LoRA‑style sub‑network inserted in parallel with FFN experts.\\
Activated parameters &
Trainable weights that are actually executed for a given token batch (depends on Top‑$K$ routing).\\
Activated‑parameter efficiency &
Ratio of activated trainable parameters to \emph{total} activated parameters in the MoE layer.\\
Characteristic hidden state $\bm h_i$ &
Prototype token embedding maximally activating $\bm g_i$ and its paired $\bm k_i$.\\
\bottomrule
\end{tabular}
\caption{Key terminology used throughout §\ref{method}–§\ref{exp}.}
\label{tab:terminology}
\end{table}
}

\subsection{The Dynamics}
\label{dynamics}

For a transformer with $L$ layers, each with attention and a Feed-Forward Network (FFN), given token embeddings $\bm x_0^{1:T}\in\mathbb R^{T\times D}$, layer $l$ computes:\footnote{
LayerNorms and dropout are omitted for clarity.} 
\begin{align}
\bm{h}_l^{1:T} &= \operatorname{SelfAttn}_{l}\!\left(\bm{x}_{l-1}^{1:T}\right) + \bm{x}_{l-1}^{1:T},
\\[2pt] 
\bm{x}_l^{t}   &= \operatorname{FFN}_{l}\!\left(\bm{h}_l^{t}\right) + \bm{h}_l^{t}.
\end{align}
\rmv{
where 
the self-attention module $\operatorname{SelfAttn}_{l}(\bm{x}_{l-1}^{1:T})$ concatenates for all attention heads their attention output across the entire token sequence from the previous hidden state $\bm{x}_{l-1}^{1:T}$.
$\bm{h}_l^{1:T}$ denotes attention output with residual connection.
}

\paragraph{Key Memory Vectors.}
A standard FFN takes form as $\sigma(\bm{h}\bm{W}_\text{up})\bm{W}_\text{down}$\footnote{
For alternative FFN structures, see Appendix \ref{GLU}.}, where $\sigma(\cdot)$ represents the activation.
Following the key-value memory perspective of \citet{geva2020transformer}, each column \(\bm{k}_i\in\mathbb{R}^D\) in \(\bm{W}_\text{up}\) serves as a key memory vector that fires on certain input patterns.
Projecting $\bm{h}^t \in \mathbb{R}^D$ onto these keys yields activation scores $\bm{a}^t \in \mathbb{R}^{D_a}$ (Figure \ref{fig:router}a). 
These key vectors function as specialized $\bm{h}^t$ pattern detectors, with their activations determining the subsequent output of the value memory vectors for each token.

\paragraph{Expert Vectors.}
Scaling up transformers brings redundancy in FFN, with most tokens trigger only a few keys \citep{elhage2021mathematical}. 
MoE groups key memory vectors into $N$ sparse experts $E_i$. A router $G(\cdot)$ picks the top‑$K$ experts per token:
\begin{align}
\operatorname{FFN}\!\bigl(\bm{h}^t\bigr)
  &= \sum_{i=1}^{N} G_i(\bm{h}^t)\,E_i(\bm{h}^t),
\\
G_i(\bm{h}^t)
  &= \operatorname{TopK}\!\Bigl(
       \operatorname{Softmax}\!\bigl(\bm{h}^t\bm{W}_g\bigr)
     \Bigr)_i.
\label{eq:moe}
\end{align}
The router learns its weight matrix $\bm{W}_g \in \mathbb{R}^{D \times N}$ that can be interpreted as a set of $N$ individual $D$-dimensional expert vectors $\bm{g}_i$, each responding to a characteristic hidden state $\bm{h}_i$ that should activate the corresponding expert $E_i$ (and their key memory vectors) \citep{zhou2022mixture}, as illustrated in Figure \ref{fig:router}b. During training, $G$ dynamically learns which $\bm{g}_i$ and ${\bm k_i}$ should better fire together.

\paragraph{PEFT for MoE.}\label{Dynamics}
A PEFT block
$
\Delta (\bm{h})=\operatorname{UpProj}\left(\operatorname{Act}\left(\operatorname{DownProj}(\bm{h})\right)\right)
    \label{eq:PEFT}
$ 
mirrors the FFN structure but is much smaller \citep{he2021towards}, with $\operatorname{Act}(\cdot)$ as non-linear $\sigma(\cdot)$ or identity function in LoRA.
Its down‑projection contains new keys $\tilde{\bm k}_j$ that respond to task‑specific patterns.

When integrating PEFT into MoE, we can choose between several intuitive approaches.
A straightforward but limited one is \textit{MoE-agnostic} adaptation of individual matrices, which fails to leverage any of the rich dynamics\footnote{As Figure \ref{fig:router}c, and discussed in §\ref{agnostic} \& Appendix \ref{addexpsetup}.} described above.
We focuses on the other approach that introduces PEFT module(s) in parallel with FFN experts\footnote{As MoE experts run in parallel and prior work shows parallel PEFT works the best \citep{he2021towards,hu2023llm,luo2024moelora,hao2024meft}, we only consider parallel composition of PEFT modules in this study.}.
This brings additional configurations with intriguing dynamics.
A single parallel PEFT module acts as a shared expert that is always active \citep{dai2024deepseekmoe}.
Alternatively, we can attach $M$ PEFT experts with their own router $\tilde G(\cdot)$ (Figure \ref{fig:router}c). The two routers, $\bm g_i$ and $\tilde{\bm g}_j$, can interact so that $\tilde{\bm k}_j$ can either refine existing subspaces or explore new ones.
This interaction can substantially enlarge the adaptation space while keeping the backbone frozen.

\rmv{
The following sections analyze how each PEFT design interacts with the underlying MoE architecture. We report its impact on performance, parameter efficiency, and the activation patterns of $\bm{k}_i$, $\bm{g}_i$, $\tilde{\bm{k}}_i$, and $\tilde{\bm{g}}_i$.
}

\subsection{The Framework}
\label{framework}

\begin{figure}[t]
\centering
\includegraphics[width=0.88\columnwidth]{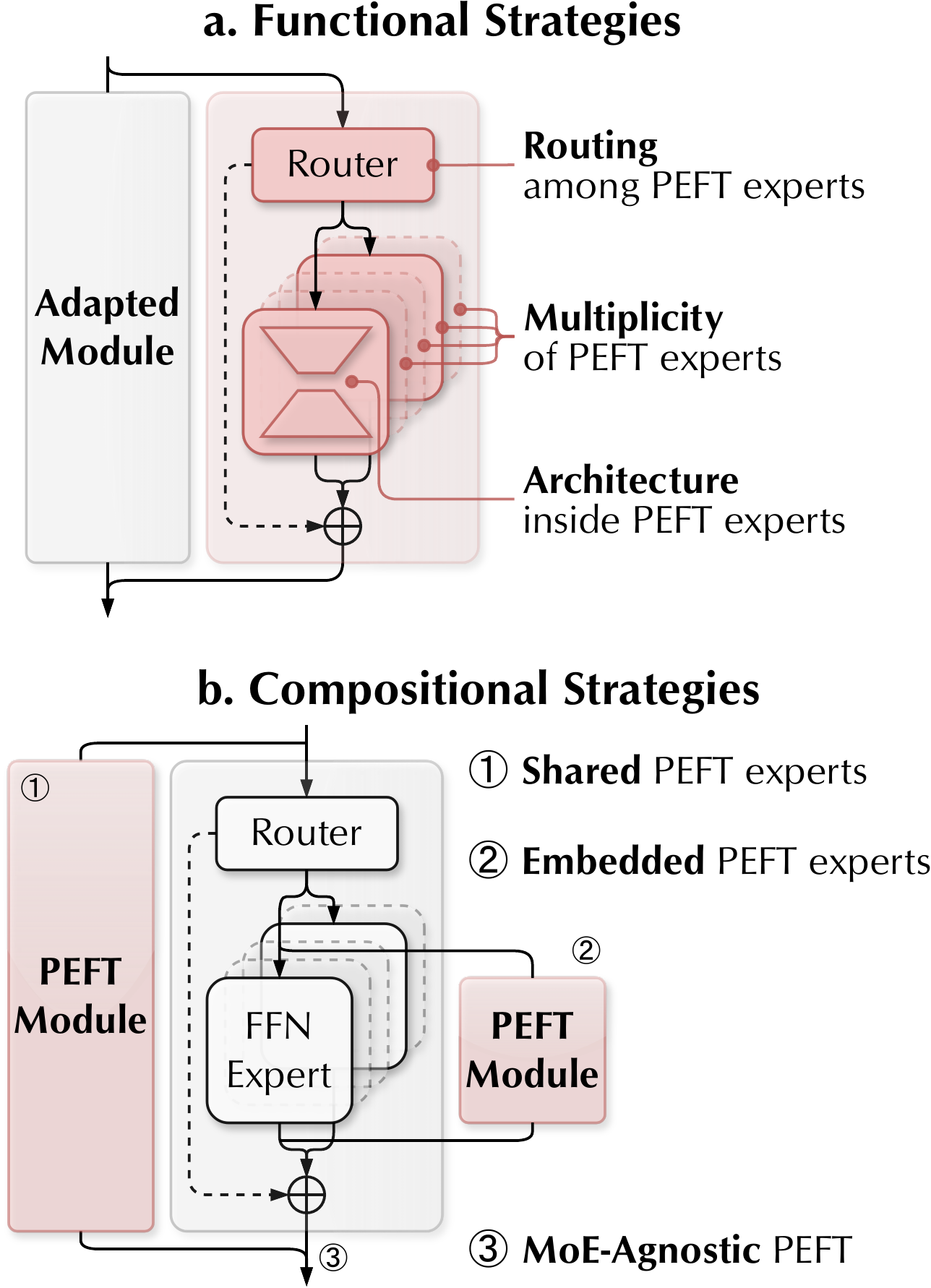}
\caption{\textbf{The framework of how PEFT designs can integrate with an MoE module.} \textbf{a.} Functional strategies specify the internal implementation of the PEFT module introduced. \textbf{b.} Compositional strategies describe the PEFT module's interaction with the original MoE mechanism.}
\label{fig:framework}
\end{figure}

Based on our insights in §\ref{dynamics}, we examine how PEFT designs can integrate with MoE. 
As illustrated in Figure \ref{fig:framework}, we introduce a framework focusing on two key design dimensions: how the adaptation modules operate, and how they interact with MoE's existing expert routing mechanisms.

\begin{figure*}[t]
    \centering
    \includegraphics[width=1\linewidth]{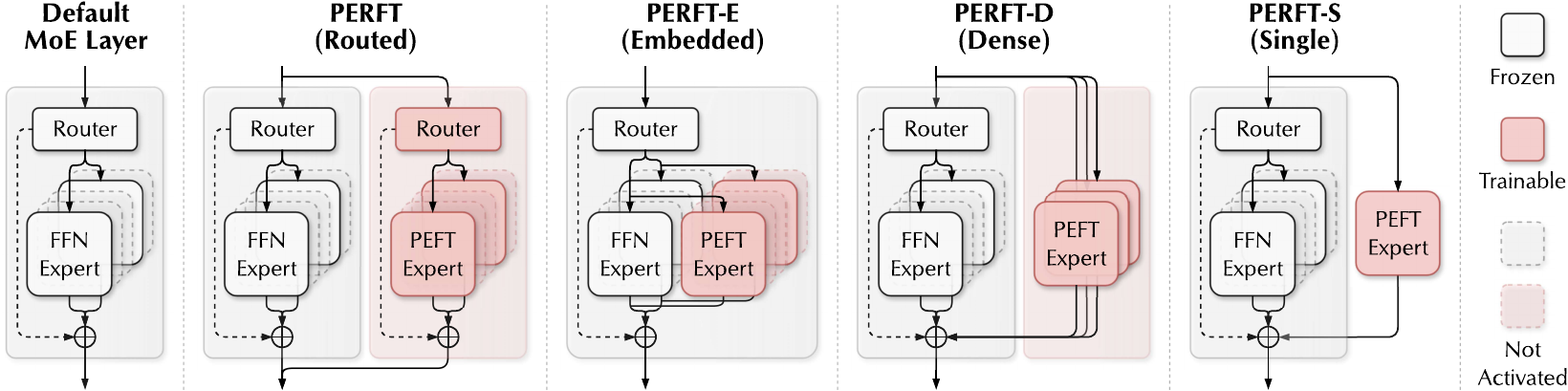}
    \caption{\textbf{Illustration of \textsc{Perft} and its ablated variants.} 
    \textsc{Perft} holds an independent routing among the introduced PEFT experts.
    \textsc{Perft}-E embeds PEFT experts within the original MoE module and directly utilizes its routing patterns. 
    \textsc{Perft}-D and \textsc{Perft}-S simply work as independent shared expert(s) alongside the MoE module.
    }
    \label{fig:architecture}
\end{figure*}

\subsubsection{Functional Strategies} 
\label{functional}

\paragraph{Architecture inside PEFT Experts.} 
Each PEFT expert uses the bottleneck layout in Eq.\ref{eq:PEFT}: $\operatorname{DownProj}(\cdot): \mathbb{R}^D\mapsto\mathbb{R}^{D_{B}}$ and $\operatorname{UpProj}(\cdot): \mathbb{R}^{D_{B}}\mapsto\mathbb{R}^D$. The bottleneck $D_B$ linearly sets the trainable‑parameter budget, like the rank $r$ in LoRA \citep{hu2021lora}. It controls the capacity for adaptation and the effectiveness of learning \citep{hu2021lora}. 

\paragraph{Multiplicity of PEFT Experts.}
More experts create multiple copies ${\Delta_i}$, increasing adaptation diversity. Studies on dense models show that adapter count strongly affects performance \citep{zadouri2023molora, liu2023moelora, dou2023loramoe, li2024mixlora}, and the optimum varies by task, model, and layer \citep{gao2024mola}.

\paragraph{Routing among PEFT Experts.} \label{routing}
The third is whether to add a separate router $\tilde G(\cdot)$. Prior MoE‑style PEFT targets dense LLMs \citep{hao2024meft, gao2024mola, wu2024mole}; our design leverages MoE-specific dynamics (§\ref{Dynamics}).
Token-wise routing over $M$ PEFT experts mirrors Eq.\ref{eq:moe}:
\begin{equation} \label{eq:routing_adapters}
\begin{aligned}
    \Delta(\bm{h}^t) &= \sum\nolimits^M_{i=1}\left(\tilde{G_i}\left(\bm{h}^t\right) 
    \Delta_i(\bm{h}^t) \right).
\end{aligned}
\end{equation}

\subsubsection{Compositional Strategies}

\paragraph{Shared PEFT Experts.} 
A single PEFT block can act as a shared expert that runs in parallel with the MoE layer. With input $\bm{h}^{1:T}$, we have:
\begin{equation}\label{eq:moe-residual}
\begin{aligned}
\bm{x}^{1:T} &=
  \sum_{i=1}^{N} G_i\!\bigl(\bm{h}^{1:T}\bigr)\,
                   E_i\!\bigl(\bm{h}^{1:T}\bigr)
\\
&\quad + \Delta\!\bigl(\bm{h}^{1:T}\bigr)
      + \bm{h}^{1:T}.
\end{aligned}
\end{equation}
The PEFT block sees the same input and adds its output to the residual stream alongside the MoE result.
Like shared FFN experts, this block captures common adaptations for all routed experts and can raise parameter efficiency.

\paragraph{Embedded PEFT Experts.} 
\label{embedded}
Here, each PEFT expert pairs with one FFN expert and receives the same token‑wise input from the MoE router:
\begin{equation}\label{eq:embedded}
\bm x^{t} =
\sum_{i=1}^{N}\! G_i(\bm h^{t})\bigl(E_i(\bm h^{t})+\Delta_i(\bm h^{t})\bigr)
+ \bm h^{t},
\end{equation}
where both outputs are weighted by $G_i$ and then added to the residual.

\paragraph{MoE-Agnostic PEFT.} \label{agnostic}
MoE‑agnostic PEFT treats the model as dense and ignores routing mechanisms. We keep it as a baseline to compare the gains of our MoE-aware designs.

\subsection{The Strategies}
\label{strategies}

Within our framework of all meaningful design choices, we implement Parameter-Efficient Routed Fine-Tuning (\textsc{Perft}), a PEFT strategy tailored for MoE models (Figure \ref{fig:architecture}), whose parallel block owns an independent router:
\begin{equation}\label{eq:moe-peft}
\begin{aligned}
\bm{x}^{1:T} &=
  \sum_{i=1}^{N} G_i\!\bigl(\bm{h}^{1:T}\bigr)\,E_i\!\bigl(\bm{h}^{1:T}\bigr)
\\
&\quad + \sum_{j=1}^{M} \tilde G_j\!\bigl(\bm{h}^{1:T}\bigr)\,
        \Delta_j\!\bigl(\bm{h}^{1:T}\bigr)
      + \bm{h}^{1:T}.
\end{aligned}
\end{equation}
The new $\tilde{G}(\cdot): \mathbb{R}^D \mapsto \mathbb{R}^M$ introduces vectors $\tilde{\bm g}_j$ that interact with $\bm g_i$ and enable flexible adaptation, as demonstrated in §\ref{routing} and Figure \ref{fig:router}c.

If $M$ equals $N$, we can also reuse the pretrained $G$ and yield the variant \textbf{\textsc{Perft}-E (Embedded)}:
\begin{equation}\label{eq:moe-residual-compact}
\begin{aligned}
\bm{x}^{1:T} &=
  \sum_{i=1}^{N} G_i(\bm{h}^{1:T})\,E_i(\bm{h}^{1:T})
\\
&\quad + \sum_{i=1}^{N} G_i(\bm{h}^{1:T})\,\Delta_i(\bm{h}^{1:T})
      + \bm{h}^{1:T}
\\[2pt]
&= \sum_{i=1}^{N} G_i(\bm{h}^{1:T})
     \bigl(E_i+\Delta_i\bigr)\!(\bm{h}^{1:T})
   + \bm{h}^{1:T}.
\end{aligned}
\end{equation}
Our experiments show that reusing $G$ helps when data are too scarce to train a fresh router.

Dropping the routing mechanism and sharing all PEFT experts gives \textbf{\textsc{Perft}-D (Dense):}
\begin{equation}\label{eq:moe-two-sums}
\begin{aligned}
\bm{x}^{1:T} &=
  \sum_{i=1}^{N} G_i(\bm{h}^{1:T})\,E_i(\bm{h}^{1:T})
\\
&\quad + \sum_{j=1}^{M} \Delta_j(\bm{h}^{1:T})
      + \bm{h}^{1:T}.
\end{aligned}
\end{equation}
And further collapsing the $M$ blocks into one yields \textbf{\textsc{Perft}-S (Single):}
\begin{equation}\label{eq:moe-one-delta}
\begin{aligned}
\bm{x}^{1:T} &=
  \sum_{i=1}^{N} G_i(\bm{h}^{1:T})\,E_i(\bm{h}^{1:T})
\\
&\quad + \Delta_0(\bm{h}^{1:T})
      + \bm{h}^{1:T}.
\end{aligned}
\end{equation}

Together, with \textsc{Perft} and its ablated variants, we can systematically experiment the design choices in our framework and verify if MoE demands a mixture of adaptation modules as expected.
\section{Experiments and Analyses}
\label{exp}

\subsection{Experiment Setup}
\label{expsetup}

\textbf{Datasets.} 
We follow the benchmark suite proposed by \citet{hu2023llm}. It contains 8 commonsense-reasoning datasets and 6 arithmetic-reasoning datasets.  
We utilize their amalgamated training sets Commonsense170K and Math50K to fine-tune models respectively for each domain. Evaluations are conducted correspondingly across all individual benchmark test sets.

\textbf{LLM Backbones.}
We use two open-source MoE LLMs as backbones: \olmoe{} \citep{muennighoff2024olmoe} and \mixtral{} \citep{jiang2024mixtral}, selected among publicly available MoE models based on their outstanding performance in the 1B and 10B activated parameter ranges. 

\textbf{Baselines.}
Applying LoRA to attention matrices $\bm{W}_q$ and $\bm{W}_v$ is the most popular PEFT setting under a tight parameter budget \citep{hu2021lora}. We therefore adopt it as our primary baseline for all scales and tasks. For the smaller \olmoe{}, we additionally LoRA-tune the router matrix $\bm{W}_g$ (results in Table \ref{tab:result-commonsense-olmoe-1}, Appendix \ref{appendix-results}).

Additional training details and design choices are provided in Appendix \ref{addexpsetup}.

\begin{figure*}[t]
  \centering

  \begin{subfigure}[b]{0.95\textwidth}
    \centering
    \includegraphics[width=\textwidth]{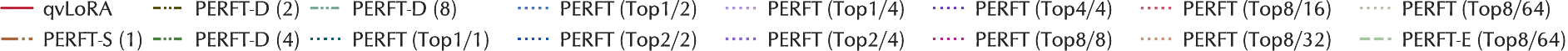}
  \end{subfigure}\par\medskip

  \begin{subfigure}[b]{0.56\textwidth}
    \centering
    \includegraphics[width=\textwidth]{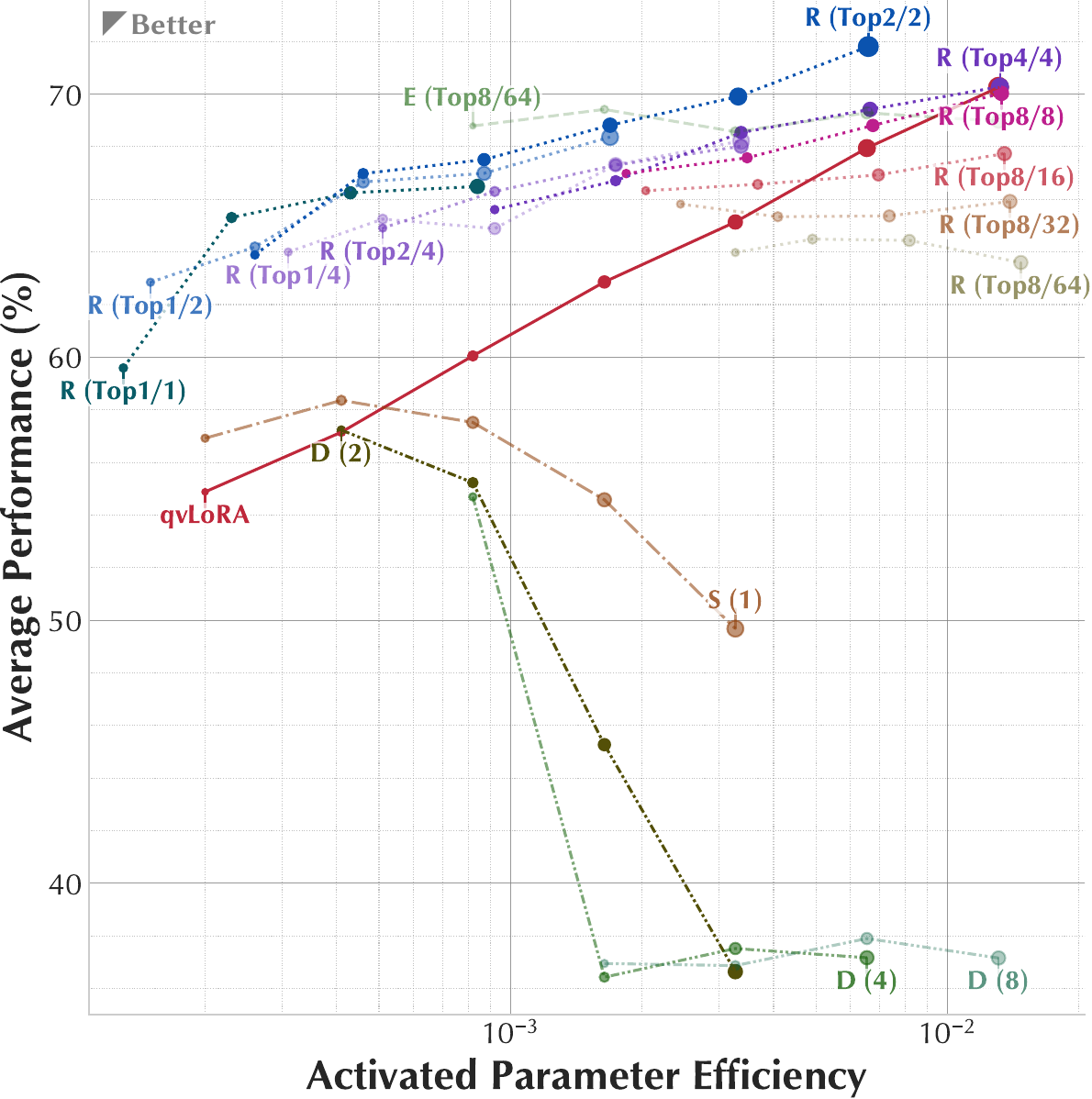}
    \caption{Commonsense Reasoning}
    \label{fig:olmoe-commonsense}
  \end{subfigure}
  \hfill
  \begin{subfigure}[b]{0.42\textwidth}
    \centering
    \includegraphics[width=\textwidth]{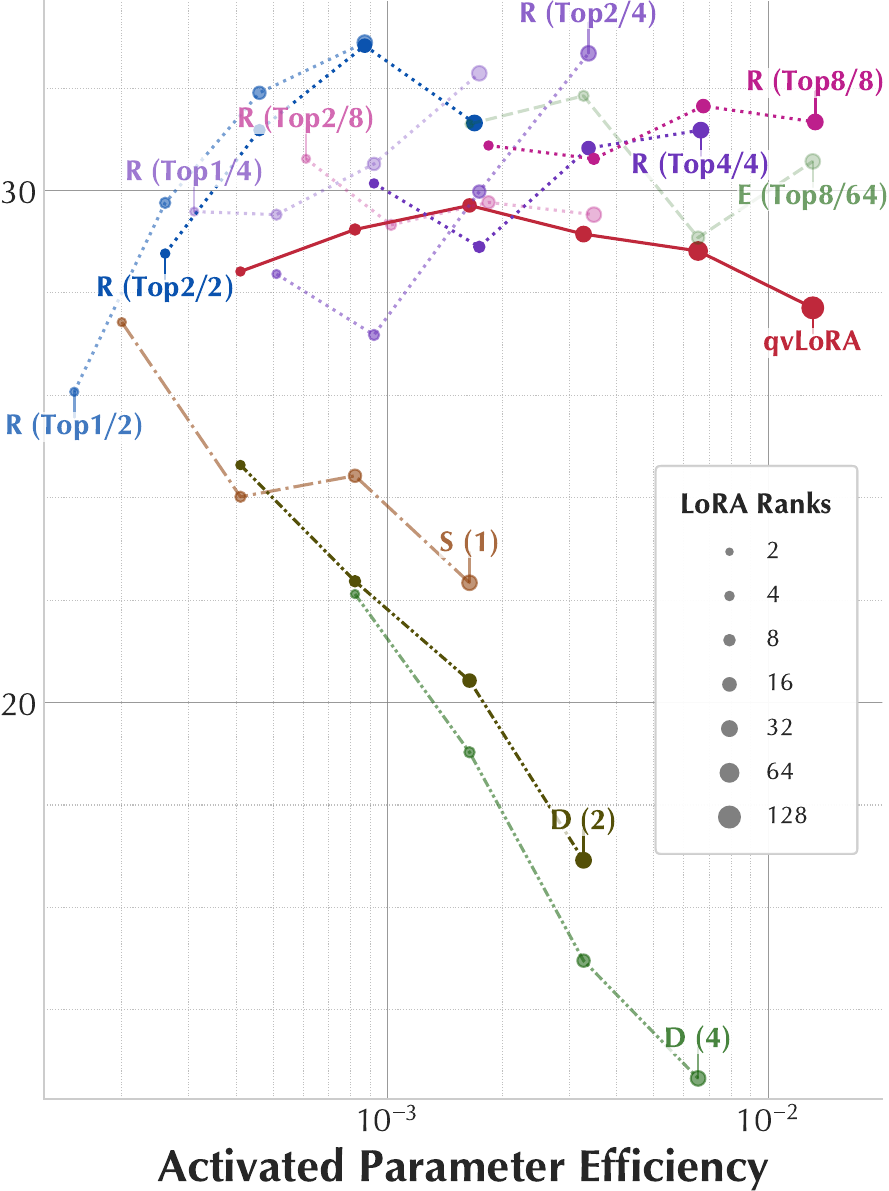}
    \caption{Arithmetic Reasoning}
    \label{fig:olmoe-math}
  \end{subfigure}

    \caption{\textbf{Performance of OLMoE fine-tuned with baselines and \textsc{Perft}.}  Scores on $y$-axes are averaged performance across each individual benchmark; Activated Parameter Efficiency on $x$-axes indicates the ratio of activated trainable parameters to the total activated parameters. 
    ``qvLoRA'' stands for applying LoRA on attention matrices $\bm{W}_q$ and $\bm{W}_v$.
    Transparency indicates different sparsity levels (ratio of activated PEFT experts).}
  \label{fig:olmoe}
\end{figure*}

\begin{table}[t]
\centering
\resizebox{\linewidth}{!}{%
\begin{tabular}{c|ll|cc|cc}
\toprule
    \textbf{LLM}  & \textbf{Arch.} & \textbf{Strategy}
    & \textbf{\# Act.} & \textbf{\% Act.}
    & \textbf{CR8} & \textbf{AR6} \\      
\midrule
    & LoRA$_{4}$  & $\bm{W}_q, \bm{W}_v$@$\texttt{Attn}$ 
    & 0.52M & 0.041 
    & 57.15 & 28.42 \\
    
    &\cc  LoRA$_{16}$ &\cc \textsc{Perft} (Top1/2)
    &\cc 0.59M &\cc 0.046
    &\cc 66.66 &\cc \textbf{31.91}\\

    &\cc  LoRA$_{8}$ &\cc \textsc{Perft} (Top2/2)
    &\cc 0.59M &\cc 0.046
    &\cc \textbf{66.98} &\cc 31.18\\
    
\cmidrule{2-7}
    \multirow{3}{*}{\shortstack[c]{OLMoE\\1B-7B\\(Top8/64)}}    
    & LoRA$_{16}$  & $\bm{W}_q, \bm{W}_v$@$\texttt{Attn}$ 
    & 2.10M & 0.164
    & 62.86 & 29.71 \\
    
    &\cc  LoRA$_{32}$ &\cc \textsc{Perft} (Top1/4)
    &\cc 2.23M &\cc 0.174
    &\cc 67.32 &\cc \textbf{32.29}\\
    &\cc  LoRA$_{4}$ &\cc \textsc{Perft}-E (Top8/64)
    &\cc 2.10M &\cc 0.164
    &\cc \textbf{69.42} &\cc 31.30\\
    
\cmidrule{2-7}
    & LoRA$_{64}$  & $\bm{W}_q, \bm{W}_v$@$\texttt{Attn}$ 
    & 8.39M & 0.654
    & 67.95 & 28.82 \\
    
    &\cc  LoRA$_{16}$ &\cc \textsc{Perft} (Top8/8)
    &\cc 8.65M &\cc 0.675
    &\cc 68.81 &\cc \textbf{31.65}\\

    &\cc  LoRA$_{16}$ &\cc \textsc{Perft}-E (Top8/64)
    &\cc 8.39M &\cc 0.654
    &\cc \textbf{69.29} &\cc 29.08 \\
\midrule
    \multirow{3}{*}{\shortstack[c]{Mixtral\\13B-47B\\(Top2/8)}}
    & LoRA$_{8}$  & $\bm{W}_q, \bm{W}_v$@$\texttt{Attn}$ 
    & 3.41M & 0.026 
    & 85.02 & 64.72 \\

    &\cc LoRA$_{8}$ &\cc \textsc{Perft} (Top2/2)
    &\cc 4.46M &\cc 0.035
    &\cc \textbf{86.23} &\cc \textbf{69.03} \\
    
    &\cc  LoRA$_{8}$ &\cc \textsc{Perft} (Top2/8)
    &\cc 5.24M &\cc 0.046
    &\cc 85.68 &\cc 68.14\\
\bottomrule
\end{tabular}%
} 
\caption{\textbf{Average performance of baseline and \textsc{Perft} variants on 8 commonsense reasoning (CR8) and 6 arithmetic reasoning (AR6) benchmarks.}
 ``Arch.'' denotes architecture inside PEFT experts. ``\#Act.'' and ``\%Act.'' represent the number of activated trainable parameters and their ratio to the total activated. ``(Top K/N)'' refers to activating $K$ of $N$ experts. Performance is the mean across individual benchmarks.}
\label{tab:result}
\end{table}

\begin{figure*}[t]
\includegraphics[width=\linewidth]{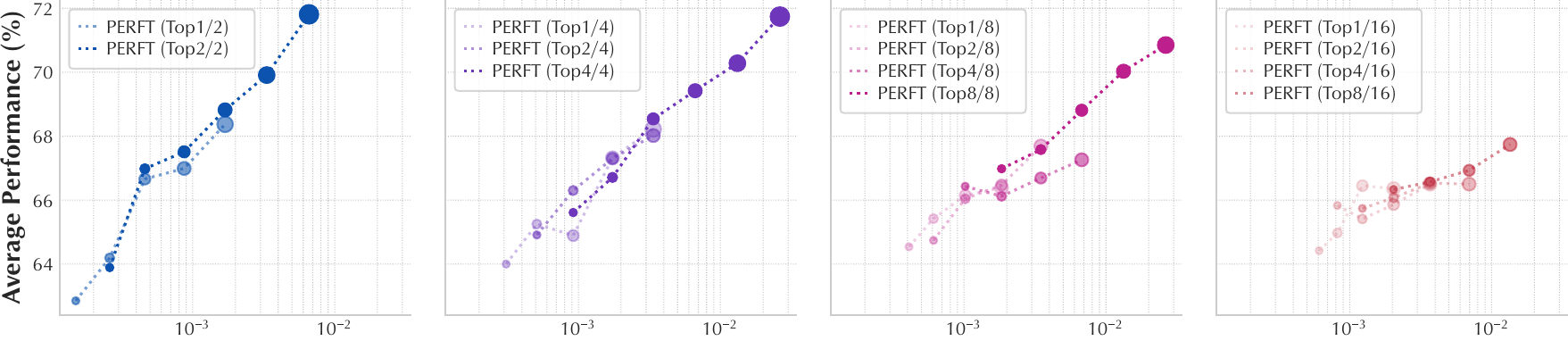}
\caption{\textbf{Performance of \textsc{Perft} configurations with different total number and activated number of PEFT experts.} Results from OLMoE fine-tuned for commonsense reasoning. $x$-axes indicate activated parameter efficiency. Transparency represents different sparsity levels. Marker size represents bottleneck size $D_B$.}
\label{fig:olmoe-commonsense-same-total}
\end{figure*}

\subsection{Experiment Results}
\label{expresult}
We validate the optimal configurations by exhaustively fine-tuning OLMoE under each configuration. The results are summarized in Figure \ref{fig:olmoe}.
Table \ref{tab:result} presents a numerical comparison between some well-performing \textsc{Perft} configurations and MoE-agnostic baselines with equivalent levels of activated trainable parameters. \textsc{Perft} improves by up to 17.2\% in commonsense and 12.3\% in arithmetic. \textsc{Perft}-E reaches 10.4\% and 5.4\%, respectively.\footnote{Notice that the reported \textsc{Perft}-E performs better than \textsc{Perft} on commonsense reasoning tasks with similar activated trainable parameters, yet this is achieved with much higher total number of trainable parameters, which is intuitive as commonsense reasoning is more knowledge-intensive and benefit from a broader pool of PEFT experts.}
Appendix \ref{appendix-results} lists full results for each configuration and task.

\textbf{\textsc{PERFT} outperforms baselines.}
Our results verified that designing PEFT with considering the underlying MoE mechanisms can indeed achieve better results.
Notably, \textsc{Perft} and its variants yields drastically different performance patterns. \textsc{Perft} and \textsc{Perft}-E are the best-performing variants, especially at higher parameter‑efficiency levels. 

\textbf{\textsc{Perft} and \textsc{Perft}-E can benefit from scaling up.}
Different variants show different scaling performances. \textsc{Perft} and \textsc{Perft}-E gain from larger bottleneck sizes $D_B$ within a certain range (shown by bigger markers in Figure \ref{fig:olmoe}). 

\textbf{\textsc{Perft} is more sensitive to overall PEFT expert number rather than activated ratio.} Figure \ref{fig:olmoe-commonsense-same-total} isolates the effect of total activated PEFT-expert count and trainable parameter efficiency. When fixing total number, the performance gain from increasing the activated ratio is relatively modest.

Additional ablations appear in Figure \ref{fig:olmoe-commonsense-routing-more} (Appendix \ref{appendix_results_olmoe}). They underline the need to balance expert count, sparsity and computational efficiency when tuning \textsc{Perft}.

\subsection{Discussion}
\label{sec:discussion}

\paragraph{Key findings.} We observe two consistent patterns across all tasks. First, \textbf{token‑wise routing} among PEFT experts (the \textsc{Perft}‑R family) drives most of the gains and enables extreme parameter efficiency. Second, when the number of PEFT experts is large, \textbf{re‑using the pretrained MoE router} (\textsc{Perft}‑E) is more stable than training a new router from scratch.
Detailed ablations, additional figures and visual analyses are provided in Appendix~\ref{app:detailed-discussion}.

\label{app:detailed-discussion}

\subsubsection{Role of Routing}
Across most tasks and budgets, the routed variant \textsc{Perft} outperforms \textsc{Perft}-S/D/E, showing that a learnable router is the main driver of \textsc{Peft} gains.
We summarize the advantage in three aspects.

\textbf{Sparse Activation.}
Figure \ref{fig:olmoe} shows that \textsc{Perft}-S/D, which always activate every PEFT block, degrade quickly as the bottleneck widens. 
This phenomenon stems from inefficient parameter utilization in always-activated shared experts. 
Section \ref{functional} shows that the bottleneck must balance capacity against learning effectiveness to reach peak performance.
\textsc{Perft} avoids this by activating only the few experts whose keys $\tilde{\bm{k}}_i$ match the token, guided by router vectors $\tilde{\bm{g}}_i$. 
Without routing, when the PEFT module's dimensions exceed the intrinsic amount required, the surplus capacity becomes detrimental rather than beneficial.

\textbf{Weight Distribution.}
When $\tilde G(\cdot)$ is absent, adding more PEFT experts hurts performance: \textsc{Perft}‑D consistently lags behind \textsc{Perft}‑S, and the gap widens as the expert count grows.
Even when every \textsc{Peft} is allowed to fire (Top$N/N$), \textsc{Perft} still beats non-routed baselines, confirming that token-wise weights, not mere capacity, lift performance.
The router assigns token‑wise gating weights, letting the model control how much each expert adapts. This dynamic weighting improves capacity utilization and supports the analysis in §\ref{routing}. 
This operates similarly to how Gated Linear Units (GLU) improve FFN layers \citep{glu}.
Without such a mechanism, the potential benefits of multiple PEFT experts would be counterbalanced by the redundancy across them.

\begin{figure*}[t]
    \centering
    \includegraphics[width=\linewidth]{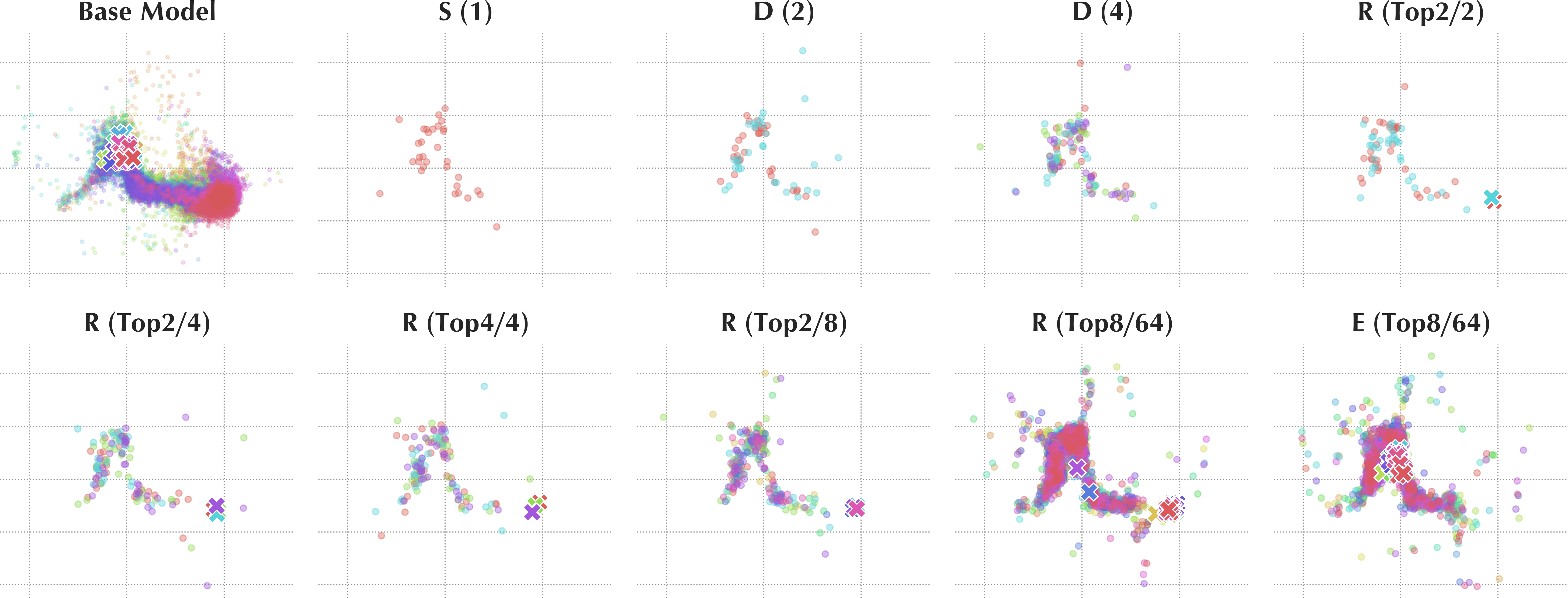}
    \caption{\textbf{Visualization of key memory vectors and expert vectors in OLMoE and \textsc{Perft} fine-tuned for commonsense reasoning.}  Results show projections of vectors with $D_B=32$ from layer 8 of OLMoE. Each subplot corresponds to a different configuration: ``Base Model'' showing vectors of FFN experts and router in the original MoE layer; ``S'', ``D'', ``R'' and ``E'' referring to vectors in the PEFT experts and router (if any) of the corresponding \textsc{Perft} variants. Markers \ding{108} represent key memory vectors in FFN or PEFT experts, and \ding{54} expert vectors in routers for either FFN experts (in Base Model and \textsc{Perft}-E) or PEFT experts (in \textsc{Perft}). All vectors are projected using the same PCA and UMAP trained on FFN experts' key memory vectors. Different colors distinguish vectors associated with different indices.}
    \label{fig:umap-commonsense-7}
\end{figure*}

\textbf{Efficiency.}
With effective routing, total PEFT capacity module matters more than the number of the activated parameters, enabling highly efficient adaptation. 
Figure \ref{fig:olmoe-commonsense-same-total} shows that for a fixed total number of PEFT experts, increasing the sparsity by activating fewer PEFT experts does not severely impact performance. 
Figure \ref{fig:umap-commonsense-7} supports this result with UMAP projections of $\bm{k}_i$ and $\bm{g}_i$ in OLMoE and $\tilde{\bm{k}}_i$ and $\tilde{\bm{g}}_i$ in different \textsc{Perft} variants.
Comparing Top2/4 with Top4/4, it confirms that an adequate subset of activated $\tilde{\bm{k}}_i$ is sufficient to capture the appropriate adaptation space.

\subsubsection{Pretrained Routing}
The relationship between \textsc{Perft}-E and \textsc{Perft} reveals important insights about leveraging pretrained knowledge versus learning new adaptation patterns, as discussed in Section \ref{embedded}.
We notice that the performance between \textsc{Perft}-E and \textsc{Perft} can vary in practice, especially when considering scenarios with different activated parameters. 
Results in Figure \ref{fig:olmoe-commonsense} show that given the same total number of PEFT experts, \textsc{Perft}-E outperforms \textsc{Perft} (Top8/64) across all bottleneck sizes; while many \textsc{Perft} configurations with fewer experts in turn outperform \textsc{Perft}-E.
Figure \ref{fig:umap-commonsense-7} illustrates the distinct dynamics between \textsc{Perft}-E and \textsc{Perft}. 
\textsc{Perft}-E utilizes the frozen $\bm{g}_i$ in $G(\cdot)$ for FFN experts, while \textsc{Perft} learns an independent $\tilde{G}(\cdot)$ from scratch for PEFT experts. 
These results suggest that when using a larger number of PEFT experts, leveraging the well-pretrained $G(\cdot)$, which already encodes effective patterns for distributing hidden space across FFN experts, would provide more stable and efficient learning for PEFT experts. 
In contrast \textsc{Perft} may expend much training resources exploring larger subspaces without effectively capturing the optimal distribution patterns for a large number of PEFT experts.
This variability highlights the complex trade-off between the flexibility offered by learning new routing mechanisms versus the stability gained from utilizing pretrained components in large-scale models, underscoring the need to consider training configuration- and task-specific factors when choosing between these approaches for large-scale model adaptation.

\rmv{
Figure \ref{fig:umap-commonsense-7} 
As the UMAP projection maintains the local distances between FFN experts' $\bm{k}_i$ in the final results, under an ideal adaptation scenario, PEFT expert's $\tilde{\bm{k}}_i$ that may activate simultaneously should ######### across task-relevant FFN experts' $\bm{k}_i$ to maximize hidden space utilization.
}

\section{Related Work}

\subsection{Mixture-of-Experts}
MoE was originally introduced as a viable solution to the computational challenges of scaling up and improving specialization \citep{jacobs1991adaptive, jordan1994hierarchical, eigen2013learning, shazeer2017outrageously}. With the rise of transformers, researchers observed that FFNs hold the largest share of parameters and capture substantial knowledge \citep{geva2020transformer, dai2021knowledge}. This capacity is linked to sparsely represented features in their activations \citep{dalvi2019one, durrani2020analyzing, gurnee2023finding}. MoE leverages this sparsity by activating only a subset of experts for each input, which improves resource utilization \citep{liu2023towards}. The idea has led to several successful MoE LLMs \citep{lepikhin2020gshard, du2022glam, fedus2022switch, zoph2022designing, jiang2024mixtral, dai2024deepseekmoe, qwen_moe, grok_1, deepseekR1}\rmv{, vision models \citep{riquelme2021scaling, liu2024task}, and multimodal models \citep{mustafa2022multimodal, shen2023scaling, lin2024moe} models. }. Recent studies explore \emph{shared experts}, modules that run in parallel with routed FFN experts and remain active for every token. This design captures common knowledge and can improve parameter efficiency \citep{gou2023mocle, dai2024deepseekmoe, qwen_moe}.
\rmv{
These advancements highlight the critical need for continued investigation into MoE models and their optimization strategies.
}

\subsection{Parameter-Efficient Fine-tuning}
Classical full fine-tuning approaches have become increasingly expensive as transformers scale \citep{devlin2018bert, qiu2020pre}.  
Recent work introduce diverse PEFT methods offering comparable performance with significantly reduced computational demands. 
\citet{he2021towards} present a unified view for PEFT, where any PEFT method can be viewed as a combination of several design dimensions. 
This perspective has inspired many hybrid designs. 
They also show that parallel PEFT modules outperform sequential ones and that modifying FFN is more effective than modifying attention. Later studies confirm these findings \citep{hu2023llm, zhang2023adaptive, dettmers2024qlora, hao2024meft}.

Recent success of MoE has sparked MoE-structured PEFT methods. Some insert mixtures of LoRA experts into the attention layers \citep{liu2023moelora, luo2024moelora}. Others place them next to dense FFNs \citep{zadouri2023molora, dou2023loramoe, page2024multi, chen2024llava-mole, hao2024meft, li2024mixlora, wu2024mole, gao2024mola}.
All these studies primarily focus on adapting dense models, which motivates us to investigate designing PEFT modules considering the underlying routing mechanisms of MoE.
Recently, \citet{wang2024let} propose expert-specialized fine-tuning as an alternative approach to PEFT, which selectively fine-tunes the most relevant experts for downstream tasks and comes closest to this research gap, although no PEFT techniques are involved and the experts weights are modified.
In our exploration of whether MoE LLMs requires mixture of adaptation modules, we directly consider introducing PEFT modules for MoE LLMs, offering more flexible and efficient solutions while preserving the original weights untouched.

\section{Conclusion}
This study addresses the gap in efficiently adapting MoE LLMs to downstream tasks.
We investigate the dynamics of core components when performing PEFT for MoE. Building on these insights, we introduce a unified framework with a comprehensive set of design dimensions.
We further propose a flexible family of PEFT strategies tailored for MoE modules. 
Extensive experiments on OLMoE and Mixtral, covering commonsense and arithmetic reasoning, show that our methods outperform MoE-agnostic baselines in both effectiveness and scalability.
We identify the optimal configuration for each design dimension and analyze the results. These observations provide practical guidance for future PEFT and MoE applications.
\section*{Limitations}
\label{sec:limitations}

\paragraph{Model scale and hardware assumptions.}
As constrained by computational resouce budgets, all experiments are conducted on \textit{OLMoE‑1B‑7B} and \textit{Mixtral‑8$\times$7B}, i.e.\ MoE backbones whose \emph{activated} parameter counts lie in the 1B– 10B range (see §\ref{expsetup}).  
It is unclear whether the \modelname{} family keeps the same efficiency–quality trade‑off on much larger models (e.g. 70 B+) or on resource constrained devices such as edge GPUs and CPUs. We also did not measure inference latency or memory footprint. Both metrics may vary with the chosen sparsity pattern. 

\paragraph{Task coverage.}
Our evaluation focuses on 14 English benchmarks: 8 commonsense‑reasoning and 6 arithmetic‑reasoning datasets (Tables \ref{tab:result}, \ref{tab:result-commonsense-olmoe-1} - \ref{tab:result-math-mixtral}).  
The gains may not transfer to language generation, code synthesis, dialogue safety, multilingual, or low‑resource scenarios. Future work should test these settings and evaluate robustness under distribution shift (e.g. adversarial or noisy inputs).

\paragraph{Hyperparameter search cost.}
Identifying the best combination of bottleneck size $D_B$, number of PEFT experts $M$, and routing sparsity $K/N$ required an extensive grid search (§\ref{expresult}, Figures \ref{fig:olmoe-math}, \ref{fig:olmoe-commonsense-same-total}).  
Once identified, a configuration generalises across tasks. However, smaller practitioners may lack the compute to reproduce the grid search. An adaptive or automated hyperparameter policy could mitigate this issue.

\paragraph{Bias and societal risk.}
Although our benchmarks are non‑dialogue and seemingly benign, both the backbone MoE LLMs and the fine‑tuning data contain demographic and geographic skews inherited from web corpora.  
We did not perform a bias or robustness audit (e.g.\ accuracy stratified by gender or language variety), nor did we evaluate privacy leakage or data memorization.  
Consequently, downstream users should apply task‑specific fairness, privacy and safety checks before deployment.

\paragraph{Environmental impact.}
Although \modelname{} reduces \emph{trainable} parameters, it still fine‑tunes multi‑billion‑parameter backbones on A100/H100 GPUs, incurring a non‑trivial carbon footprint.  
A detailed energy accounting (e.g.\ kWh per experiment) was not recorded; future work should explore greener training (mixed‑precision, progressive pruning) and life‑cycle impact reporting.

\smallskip
\noindent These limitations highlight promising directions for extending the current study and for responsibly deploying PEFT techniques on sparse MoE LLMs.
\bibliography{custom}

\appendix
\label{sec:appendix}
\section{Additional Experiment Setup and Discussions}
\label{addexpsetup}

\subsection{Training Configurations.}
\label{config}
\textbf{Hardware.} For each experiment we trained \olmoe{} on a single NVIDIA A100 GPU. \mixtral{} was trained on our 4$\times$NVIDIA H100 GPUs connected with NV-link. Both models are evaluated on NVIDIA A100 GPUs.

\textbf{Hyperparameters.} We display the hyperparameter configurations used in fine-tuning and evaluating \olmoe{} and \mixtral{} in Table \ref{tab:hyperparam}. We use the LoRA settings recommended by \citet{hu2023llm} and keep all other hyperparameters at their model-default values.

\textbf{Loss Functions.}
In our experiments, we maintain consistency with the original training process of each LLM by incorporating their respective auxiliary losses alongside the cross-entropy loss for token outputs. 
All evaluated models include a load‑balancing loss, which encourages an equal token distribution among experts \citep{shazeer2017outrageously}. \olmoe{} additionally incorporates a router z-loss to penalize large routing logits and stabilize training \citep{zoph2022st}. 
To ensure a fair comparison, we keep all auxiliary losses active during fine-tuning for baseline and all \modelname{} variants. 
For \modelname{}, we extend this approach with the load balancing loss for the PEFT expert router as well for a similar balanced distribution of tokens among PEFT experts.
Detailed hyperparameters and resource configurations for our experiments are provided in Appendix \ref{config}.

\begin{table}[t]
\small
\centering
\begin{tabular}{c|cc}
\toprule
    \textbf{Hyperparameters} & \textbf{\olmoe{}} & \textbf{\mixtral{}} \\
\midrule   
    Training precision & \multicolumn{2}{c}{BFloat16} \\
    Dropout      & \multicolumn{2}{c}{0.05} \\
    Optimizer    & \multicolumn{2}{c}{AdamW} \\
    LR           & 1e-5 & 2e-5 \\
    LR scheduler & \multicolumn{2}{c}{Linear} \\
    Batch size   & \multicolumn{2}{c}{16} \\
    Warmup steps & \multicolumn{2}{c}{100} \\
    Epochs       & \multicolumn{2}{c}{3} \\
    Auxiliary loss coef. & 0.01 & 0.02 \\
\bottomrule
\end{tabular}
\caption{\textbf{Hyperparameter configurations for \olmoe{} and \mixtral{}.}}
\label{tab:hyperparam}
\end{table}

\subsection{Gated Linear Unit} 
\label{GLU}
Modern transformers often adopt the Gated Linear Unit (GLU), which adds an element‑wise multiplicative gate after activation \citep{glu,swiglu}. Formally: $\texttt{FFN}_\text{GLU}(\bm{h})=[\sigma(\bm{h}\bm{W}_\text{up})\otimes(\bm{h}\bm{W}_\text{gate})]\bm{W}_\text{down}$. 
We focus on the matrix $\bm{W}_\text{up}$ since it directly processes $h$ and its output passes through $\sigma(\cdot)$, which controls key-memory activation. The same argument applies to both vanilla FFN and GLU. 
\rmv{This enables input-dependent information filtering, potentially leading to more nuanced and effective representations. }
\section{Additional Analyses for Design Configurations}
\subsection{Architecture inside PEFT Experts}
\textbf{LoRA Versus Parallel Adapters.}
We centre our study on LoRA adapters because they are simple yet effective. Output scaling with $\alpha$ also reduces the need to retune hyperparameters when the bottleneck size changes \citep{yang2020feature, hu2021lora}.
Motivated by results on dense models \citep{he2021towards, hu2023llm}, we also analyze \textit{parallel adapters} \citep{houlsby2019parameter, he2021towards}, which add an activation function after the bottleneck.
\label{parallel}

\begin{table*}[t]
\centering
\resizebox{\linewidth}{!}{%
\begin{tabular}{ll|cc|cccccccc|c}
\toprule
    \textbf{Arch.} & \textbf{Strategy }
    & \textbf{\# Act.} & \textbf{\% Act.}
    & \textbf{BoolQ} & \textbf{PIQA}  & \textbf{SIQA}  & \textbf{HellaS}
    & \textbf{WinoG} & \textbf{ARC-e} & \textbf{ARC-c} & \textbf{OBQA}  & \textbf{Avg.}\\  
    
\midrule
\rowcolor{LightRed}
    LoRA$_{4}$ & \modelname{} (Top1/1)
    & 0.16M & 0.013 
    & 62.48 & 75.73 & \textbf{68.17} & 25.16 & 51.07 & 76.81 & 55.72 & \textbf{61.60} & 59.59\\
    PA$_{4}$  & \modelname{} (Top1/1)
    & 0.16M & 0.013 
    & \textbf{63.09} & \textbf{76.50} & 64.94 & \textbf{31.23} & \textbf{52.72} & \textbf{77.02} & \textbf{56.31} & 55.40 & \textbf{59.65} \\
\rowcolor{LightRed}
    LoRA$_{8}$ & \modelname{} (Top1/1)
    & 0.29M & 0.023
    & 63.43 & 77.53 & \textbf{70.68} & \textbf{42.13} & \textbf{66.14} & 77.10 & \textbf{59.30} & \textbf{66.20} & \textbf{65.31} \\
    PA$_{8}$ & \modelname{} (Top1/1)
    & 0.29M & 0.023
    & \textbf{65.63} & \textbf{78.94} & 68.68 & 40.46 & 53.75 & \textbf{79.25} & 56.14 & 61.20 & 63.01 \\
\rowcolor{LightRed}
    LoRA$_{16}$ & \modelname{} (Top1/1)
    & 0.56M & 0.043 
    & 64.98 & \textbf{78.56} & \textbf{72.52} & \textbf{41.99} & \textbf{67.25} & 77.82 & 58.70 & \textbf{68.20} & \textbf{66.25} \\
    PA$_{16}$ & \modelname{} (Top1/1)
    & 0.56M & 0.043 
    & \textbf{66.61} & \textbf{78.56} & 71.34 & 41.26 & 59.75 & \textbf{78.87} & \textbf{59.30} & 66.20 & 65.24 \\
\rowcolor{LightRed}
    LoRA$_{32}$ & \modelname{} (Top1/1)
    & 1.08M & 0.084 
    & 66.36 & 78.84 & 72.36 & \textbf{42.83} & 63.38 & 78.62 & 58.36 & \textbf{71.20} & 66.49 \\
    PA$_{32}$ & \modelname{} (Top1/1)
    & 1.08M & 0.084 
    & \textbf{66.61} & \textbf{79.54} & \textbf{72.62} & 42.36 & \textbf{66.46} & \textbf{79.29} & \textbf{62.03} & 67.40 & \textbf{67.04} \\
\midrule
\rowcolor{LightRed}
    LoRA$_{4}$ & \modelname{} (Top2/2)
    & 0.33M & 0.026
    & 64.86 & 76.71 & \textbf{69.60} & 40.89 & \textbf{62.43} & 77.23 & 55.80 & \textbf{63.60} & \textbf{63.89} \\
    PA$_{4}$ & \modelname{} (Top2/2)
    & 0.33M & 0.026
    & \textbf{65.44} & \textbf{77.48} & 69.40 & \textbf{41.14} & 51.54 & \textbf{78.83} & \textbf{57.94} & 63.20 & 63.12 \\
\rowcolor{LightRed}
    LoRA$_{8}$ & \modelname{} (Top2/2)
    & 0.59M & 0.046
    & 65.26 & 78.18 & \textbf{72.31} & \textbf{42.11} & \textbf{71.82} & 77.90 & \textbf{60.49} & \textbf{67.80} & \textbf{66.98} \\
    PA$_{8}$ & \modelname{} (Top2/2)
    & 0.59M & 0.046
    & \textbf{67.31} & \textbf{80.03} & 71.14 & 41.70 & 61.80 & \textbf{78.58} & 58.87 & 66.60 & 65.75 \\
\rowcolor{LightRed}
    LoRA$_{16}$ & \modelname{} (Top2/2)
    & 1.11M & 0.087 
    & 66.18 & 77.97 & \textbf{72.52} & \textbf{43.99} & \textbf{70.64} & 78.24 & 60.75 & 69.80 & 67.51 \\
    PA$_{16}$ & \modelname{} (Top2/2)
    & 1.11M & 0.087 
    & \textbf{66.76} & \textbf{79.38} & 72.47 & 43.52 & 69.85 & \textbf{80.85} & \textbf{61.26} & \textbf{71.00} & \textbf{68.14} \\
\rowcolor{LightRed}
    LoRA$_{32}$ & \modelname{} (Top2/2)
    & 2.16M & 0.169 
    & 65.81 & 79.38 & \textbf{73.59} & \textbf{49.42} & \textbf{71.59} & 77.78 & \textbf{61.18} & 71.80 & 68.82 \\
    PA$_{32}$ & \modelname{} (Top2/2)
    & 2.16M & 0.169 
    & \textbf{67.61} & \textbf{80.96} & 73.18 & 45.57 & 70.64 & \textbf{80.68} & \textbf{61.18} & \textbf{72.00} & \textbf{68.98} \\
\midrule
\rowcolor{LightRed}
    LoRA$_{4}$ & \modelname{} (Top2/4)
    & 0.66M & 0.051
    & 63.98 & 75.68 & 69.29 & 40.26 & \textbf{65.75} & 77.36 & \textbf{59.56} & \textbf{67.40} & \textbf{64.91} \\
    PA$_{4}$ & \modelname{} (Top2/4)
    & 0.66M & 0.051
    & \textbf{65.93} & \textbf{77.75} & \textbf{69.96} & \textbf{40.81} & 61.09 & \textbf{79.17} & 58.28 & 65.80 & 64.85 \\
\rowcolor{LightRed}
    LoRA$_{8}$ & \modelname{} (Top2/4)
    & 1.18M & 0.092
    & \textbf{65.02} & 77.86 & \textbf{71.90} & 41.61 & 68.75 & 77.31 & 59.13 & \textbf{68.80} & 66.30 \\
    PA$_{8}$ & \modelname{} (Top2/4)
    & 1.18M & 0.092
    & 64.40 & \textbf{78.07} & 71.24 & \textbf{41.80} & \textbf{70.17} & \textbf{79.76} & \textbf{61.09} & 67.80 & \textbf{66.79} \\
\rowcolor{LightRed}
    LoRA$_{16}$ & \modelname{} (Top2/4)
    & 2.23M & 0.174
    & 64.07 & 76.61 & \textbf{73.59} & 42.10 & \textbf{71.90} & 78.32 & \textbf{60.58} & \textbf{71.20} & \textbf{67.30} \\
    PA$_{16}$ & \modelname{} (Top2/4)
    & 2.23M & 0.174
    & \textbf{65.99} & \textbf{79.92} & 72.62 & \textbf{43.14} & 61.64 & \textbf{80.09} & \textbf{60.58} & 69.20 & 66.65 \\
\rowcolor{LightRed}
    LoRA$_{32}$ & \modelname{} (Top2/4)
    & 4.33M & 0.337
    & 66.30 & 77.75 & \textbf{75.44} & \textbf{45.88} & \textbf{71.43} & 76.18 & 60.58 & \textbf{70.60} & 68.02 \\
    PA$_{32}$ & \modelname{} (Top2/4)
    & 4.33M & 0.337
    & \textbf{66.70} & \textbf{79.33} & 73.18 & 42.57 & 70.40 & \textbf{81.10} & \textbf{62.20} & \textbf{70.60} & \textbf{68.26} \\
\bottomrule
\end{tabular}
}
\caption{\textbf{Commonsense reasoning performance of OLMoE with \modelname{} using LoRA and Parallel Adapter (PA).} ``Arch.'' denotes the architecture inside PEFT modules. ``\# Act.'' and ``\% Act.'' represent the number of activated trainable parameters and their ratio to the total activated parameters. ``(TopK/N)'' refers to activating $K$ experts among the total number of $N$ experts. Dataset names are partially abbreviated, including BoolQ \citep{clark2019boolq}, PIQA \citep{bisk2020piqa}, Social IQa \citep{sap2019siqa}, HellaSwag \citep{zellers2019hellaswag}, WinoGrande \citep{sakaguchi2021winogrande}, Easy Set and Challenge Set of ARC \citep{clark2018arc}, and OpenBookQA \citep{mihaylov2018obqa}.
}
\label{tab:parallel}
\end{table*}

Table \ref{tab:parallel} compares the commonsense reasoning performance of LoRA and Parallel Adapters (PA) as PEFT experts in \olmoe{} with several well-performing \modelname{} configurations.
As we can see, under equivalent activated trainable parameter levels, the average performance difference between LoRA and PA is only marginal.
Interestingly, on specific tasks, certain architectures consistently outperform others. 
For instance, parallel adapters generally perform better on BoolQ, PIQA, and ARC, while LoRA excels in SIQA and OBQA. 
These task‑specific gaps may reflect differences in required knowledge or data distribution. A deeper investigation into these task‑specific variations is beyond the scope of this study.
Given the similar average performance, we opted to focus on LoRA for our experiments due to its simpler structure without the additional activation function.

It is also viable to consider copying the original FFN structure as PEFT experts.
We have opted not to investigate this option further in our current study based on two reasons.
First, copying the full FFN violates the spirit of PEFT because it effectively upsizes the model to a version with more experts.
Second, recent advancements have introduced more complex implementations that go beyond the simple $\sigma(\bm{h}\bm{W}_\text{up})\bm{W}_\text{down}$ pattern how FFN initially designed as. 
GLU has become widely adopted in modern transformers including \olmoe{} and \mixtral{}.
\rmv{This enables input-dependent information filtering, potentially leading to more nuanced and effective representations. } 
The increased complexity of GLU, with its three matrices, presents challenges for a fair controlled comparison under the same parameter budget. 
Given these considerations, we focus on experimenting within our current scope.

\textbf{Bottleneck Sizes.}
We experiment with different bottleneck sizes ranging from 2 to 128. 
Here we provide a detailed empirical analysis about the inefficient parameter utilization when always-activated shared experts are employed without an effective routing mechanism. 
Such cases reveal a mismatch between task dimensionality and adapter capacity. When the bottleneck is too wide, the extra dimensions add little signal and can even hurt performance.
Large, randomly‑initialized bottlenecks in \modelname{}‑S or \modelname{}‑D inject noise into otherwise unused subspaces and may corrupt pretrained representations.
If the residual stream is viewed as limited bandwidth between modules \citep{elhage2021mathematical}, then only a small subspace should carry task‑specific adaptation when most weights stay frozen.
Any over-parameterized adaptation can unnecessarily disrupt normal functioning on the residual stream's bandwidths, potentially destabilizing the original gradient flow in the transformer and leading to unstable training or sub-optimal solutions \citep{aghajanyan2021intrinsic}.
Simultaneously, in the PEFT context with limited adaptation information compared to model pretraining, an excessively large parameter space without gating control can easily result in over-fitting on fine-tuning data, which is exacerbated by the sparse nature of the MoE module we are adapting.
As the MoE module hosts multiple different patterns on various combinations of activated FFN experts that dynamically interact with each other on the residual stream, the always-activated \modelname{}-S and \modelname{}-D variants may learn unnecessary adaptations during the training process, further aggravating the disrupted functionality and over-fitting problems. 

It is also worth noting that since FFN tends to learn task-specific textual patterns \citep{geva2020transformer} and attention learns more about positional interactions \citep{elhage2021mathematical}, the nature of different components to which PEFT is introduced also contributes to different phenomena.
For the baseline LoRA operating on attention matrices, individual attention heads are already operating on relatively smaller subspaces and can easily write outputs to disjoint subspaces without interaction. 
Because each attention head operates in a low‑rank subspace, its read/write patterns are relatively fixed.
Consequently, additional parameters introduced by scaling the bottleneck of attention LoRA  may not interfere with information from other components as severely as adapting the MoE FFN module.

\subsection{Multiplicity of PEFT Experts} 
We vary the total number of PEFT experts from 1 to 64 and the number of activated experts from 1 to 8. This grid lets us study how expert count and activation ratio affect performance. 
We denote $K$ out of $M$ routed PEFT experts activated per token as  "(TopK/M)", and $N$ shared PEFT experts without routing as "(N)".

\subsection{Routing among PEFT Experts} 
We investigate both learned routing (PERFT) and embedded routing using the pretrained MoE router (PERFT-E). We also include non-routed variants (PERFT-D/S) for comparison.
This allows us to systematically study the impact of parameter efficiency on performance across \modelname{} variants.
\newpage
\onecolumn
\section{Additional Results}
\label{appendix-results}
\subsection{\olmoe{} for Commonsense Reasoning}
\label{appendix_results_olmoe}
\begin{figure}[htbp]
\centering
\includegraphics[width=0.9\linewidth]{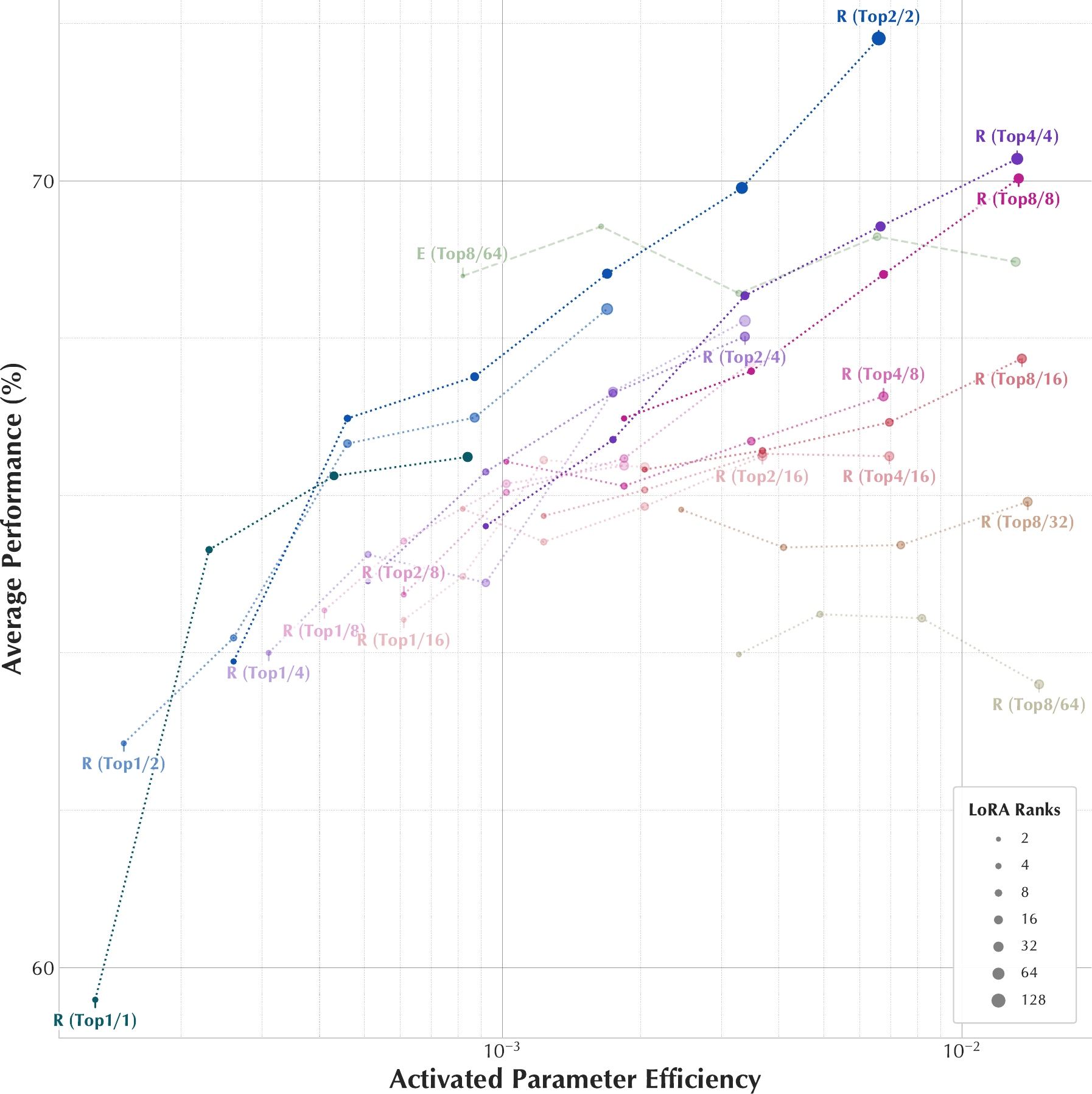}
    \caption{\textbf{Performance comparison of \olmoe{} fine-tuned with different configurations of \modelname{}.}  Performance on $y$-axes is averaged across commonsense reasoning benchmarks; ``Activated Parameter Efficiency'' on $x$-axes indicates the ratio of activated trainable parameters to the total activated parameters. 
    Color represents different configurations of \modelname{}.
    Transparency indicates different sparsity levels (ratio of activated experts $K/N$, as ``(TopK/N)'' labeled for \modelname{} and \modelname{}-E). Marker size indicates bottleneck size $D_B$.}
\label{fig:olmoe-commonsense-R}
\end{figure}

\begin{figure}[htbp]
\centering
\begin{subfigure}[b]{0.99\textwidth}
    \centering
    \includegraphics[width=\textwidth]{img/olmoe-commonsense-same-total-nox.pdf}
    \caption{Dynamics of configurations with different numbers of total PEFT experts in PERFT}
    \vspace{0.1in}
\end{subfigure}
\begin{subfigure}[b]{0.99\textwidth}
    \centering
    \includegraphics[width=\textwidth]{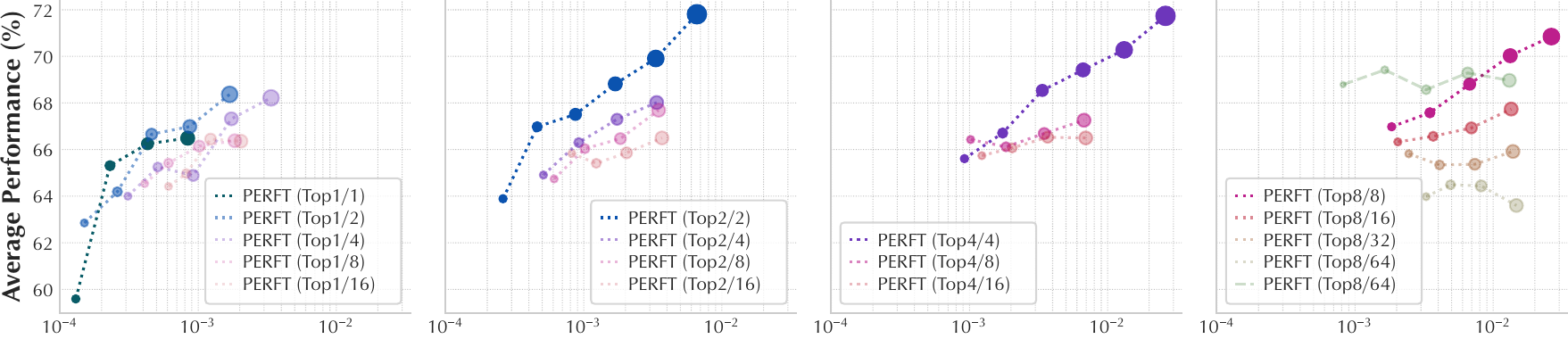}
    \caption{Dynamics of configurations with different numbers of total PEFT experts in PERFT}
    \vspace{0.1in}
\end{subfigure}
\begin{subfigure}[b]{0.99\textwidth}
    \centering
    \includegraphics[width=\textwidth]{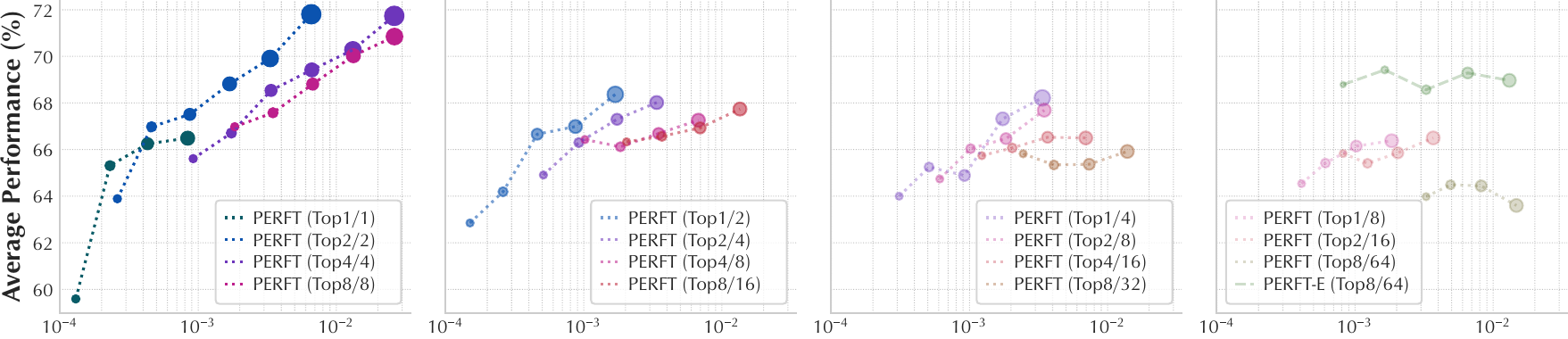}
    \caption{Dynamics of configurations with different activated ratios among PEFT experts in PERFT}
\end{subfigure}
\caption{
\textbf{Performance comparison of configurations with different total number of PEFT experts in PERFT.} Results from \olmoe{} fine-tuned with \modelname{} for commonsense reasoning. $x$-axes stand for activated parameter efficiency. Transparency represents different sparsity levels (ratio of activated PEFT experts), and marker size represents bottleneck size $D_B$.}
\label{fig:olmoe-commonsense-routing-more}
\end{figure}

\newpage

\setlength\LTcapwidth{\textwidth}
{\renewcommand{\arraystretch}{0.9}
\centering
\scriptsize
\addtolength{\tabcolsep}{-3pt} 
\begin{longtable}{ll|cc|cccccccc|c}
\toprule
    \textbf{Arch.} & \textbf{Strategy }
    & \textbf{\# Act.} & \textbf{\% Act.}
    & \textbf{BoolQ} & \textbf{PIQA}  & \textbf{SIQA}  & \textbf{HellaS}
    & \textbf{WinoG} & \textbf{ARC-e} & \textbf{ARC-c} & \textbf{OBQA}  & \textbf{Avg.}\\  
\midrule

    Base & (pretrained)
    & --- & ---
    & 42.42 & 52.61 & 16.53 & 21.27 & 28.10 & 13.13 & 13.99 & 6.80 & 24.36\\
    Base & (instruct)
    & --- & ---
    & 59.94 & 62.68 & 12.03 & 22.27 & 5.84 & 15.15 & 17.15 & 8.00 & 25.38 \\
\midrule
    LoRA$_{2}$  & $\bm{W}_q, \bm{W}_v$@$\texttt{Attn}$ 
    & 0.26M & 0.020 
    & 62.02 & 71.11 & 59.77 & 28.48 & 50.36 & 70.37 & 48.89 & 48.00 & 54.88
    \\
    LoRA$_{4}$  & $\bm{W}_q, \bm{W}_v$@$\texttt{Attn}$ 
    & 0.52M & 0.041 
    & 60.40 & 73.61 & 62.90 & 32.08 & 50.20 & 74.12 & 52.65 & 51.20 & 57.15 \\    
    LoRA$_{8}$  & $\bm{W}_q, \bm{W}_v$@$\texttt{Attn}$ 
    & 1.05M & 0.082 
    & 63.76 & 74.86 & 65.30 & 37.01 & 50.83 & 76.81 & 55.46 & 56.40 & 60.05 \\    
    LoRA$_{16}$ & $\bm{W}_q, \bm{W}_v$@$\texttt{Attn}$ 
    & 2.10M & 0.164 
    & 64.95 & 76.88 & 69.60 & 39.27 & 53.35 & 78.07 & 57.34 & 63.40 & 62.86 \\    
    LoRA$_{32}$ & $\bm{W}_q, \bm{W}_v$@$\texttt{Attn}$ 
    & 4.19M & 0.327 
    & 66.79 & 78.56 & 70.93 & 41.63 & 58.41 & 79.38 & 60.41 & 65.00 & 65.14 \\    
    LoRA$_{64}$ & $\bm{W}_q, \bm{W}_v$@$\texttt{Attn}$
    & 8.39M & 0.654
    & 67.13 & 80.30 & 73.34 & 44.28 & 65.90 & 80.72 & 61.95 & 70.00 & 67.95
    \\    
    LoRA$_{128}$ & $\bm{W}_q, \bm{W}_v$@$\texttt{Attn}$
    & 16.8M & 1.309
    & 68.32 & 82.64 & 74.16 & 45.71 & 72.45 & 81.36 & 63.82 & 73.60 & 70.26 
    \\    
\midrule
    LoRA$_{4}$  & $\bm{W}_g$@$\texttt{Gate}$ 
    & 0.14M & 0.011 
    & 62.14 & 59.79 & 39.66 & 25.94 & 51.62 & 42.63 & 36.52 & 29.00 & 43.41 \\
    LoRA$_{8}$  & $\bm{W}_g$@$\texttt{Gate}$ 
    & 0.27M & 0.021 
    & 59.11 & 66.49 & 47.59 & 27.37 & 51.70 & 52.06 & 42.06 & 33.20 & 47.45 \\
    LoRA$_{16}$  & $\bm{W}_g$@$\texttt{Gate}$ 
    & 0.54M & 0.042 
    & 62.05 & 64.04 & 47.85 & 28.08 & 49.33 & 57.37 & 43.17 & 34.40 & 48.29 \\
    LoRA$_{32}$  & $\bm{W}_g$@$\texttt{Gate}$ 
    & 1.08M & 0.084 
    & 59.24 & 60.07 & 43.19 & 26.62 & 49.09 & 41.50 & 32.34 & 31.60 & 42.96 \\
\midrule
\midrule
    LoRA$_{4}$ & \modelname{}-S (1)
    & 0.26M & 0.020
    & 63.82 & 72.31 & 63.87 & 25.45 & 50.12 & 73.91 & 49.49 & 56.40 & 56.92 \\
    LoRA$_{8}$ & \modelname{}-S (1)
    & 0.52M & 0.041
    & 63.52 & 73.56 & 66.33 & 25.45 & 51.93 & 72.60 & 52.47 & 61.00 & 58.36 \\
    LoRA$_{16}$ & \modelname{}-S (1)
    & 1.05M & 0.082
    & 63.49 & 71.71 & 65.71 & 25.11 & 51.22 & 71.13 & 50.60 & 61.20 & 57.52 \\
    LoRA$_{32}$ & \modelname{}-S (1)
    & 2.10M & 0.164
    & 62.08 & 68.28 & 64.69 & 25.37 & 52.17 & 64.73 & 44.54 & 54.80 & 54.58 \\
    LoRA$_{64}$ & \modelname{}-S (1)
    & 4.19M & 0.327
    & 61.59 & 63.76 & 59.11 & 24.48 & 54.06 & 53.75 & 36.86 & 43.80 & 49.68 \\
\midrule
\midrule
    LoRA$_{4}$ & \modelname{}-D (2)
    & 0.52M & 0.041
    & 62.14 & 71.87 & 66.53 & 25.41 & 51.07 & 72.60 & 50.43 & 57.80 & 57.23 \\
    LoRA$_{8}$ & \modelname{}-D (2)
    & 1.05M & 0.082
    & 62.87 & 71.44 & 63.41 & 25.47 & 51.70 & 65.28 & 46.84 & 54.80 & 55.23 \\
    LoRA$_{16}$ & \modelname{}-D (2)
    & 2.10M & 0.164
    & 62.14 & 59.68 & 46.98 & 25.51 & 49.25 & 45.96 & 33.45 & 39.20 & 45.27 \\
    LoRA$_{32}$ & \modelname{}-D (2)
    & 4.19M & 0.327
    & 62.17 & 48.20 & 32.86 & 25.38 & 48.86 & 24.87 & 25.17 & 25.60 & 36.64 \\
\midrule
    LoRA$_{4}$ & \modelname{}-D (4)
    & 1.05M & 0.082
    & 62.87 & 69.37 & 61.98 & 24.93 & 50.91 & 65.78 & 46.08 & 55.60 & 54.69 \\
    LoRA$_{8}$ & \modelname{}-D (4)
    & 2.10M & 0.164
    & 62.17 & 49.29 & 33.06 & 24.57 & 49.57 & 25.46 & 25.09 & 22.20 & 36.43 \\
    LoRA$_{16}$ & \modelname{}-D (4)
    & 4.19M & 0.327
    & 62.17 & 50.60 & 33.21 & 24.67 & 48.78 & 26.01 & 24.74 & 30.00 & 37.52 \\
    LoRA$_{32}$ & \modelname{}-D (4)
    & 8.39M & 0.654
    & 62.17 & 52.18 & 33.47 & 25.02 & 50.51 & 25.80 & 22.18 & 26.00 & 37.17 \\
\midrule
    LoRA$_{4}$ & \modelname{}-D (8)
    & 2.10M & 0.164
    & 62.11 & 48.86 & 35.11 & 24.57 & 48.22 & 25.51 & 23.38 & 27.80 & 36.94 \\
    LoRA$_{8}$ & \modelname{}-D (8)
    & 4.19M & 0.327
    & 62.17 & 49.13 & 33.27 & 25.37 & 49.41 & 25.00 & 24.23 & 26.40 & 36.87 \\
    LoRA$_{16}$ & \modelname{}-D (8)
    & 8.39M & 0.654
    & 62.17 & 52.01 & 33.47 & 24.91 & 53.20 & 25.29 & 26.96 & 25.20 & 37.90 \\
    LoRA$_{32}$ & \modelname{}-D (8)
    & 16.8M & 1.309
    & 62.17 & 50.92 & 33.88 & 24.58 & 49.64 & 24.16 & 26.71 & 25.20 & 37.16 \\
\midrule
\midrule
    LoRA$_{4}$ & \modelname{} (Top1/1)
    & 0.16M & 0.013
    & 62.48 & 75.73 & 68.17 & 25.16 & 51.07 & 76.81 & 55.72 & 61.60 & 59.59 \\
    LoRA$_{8}$ & \modelname{} (Top1/1)
    & 0.29M & 0.023
    & 63.43 & 77.53 & 70.68 & 42.13 & 66.14 & 77.10 & 59.30 & 66.20 & 65.31 \\
    LoRA$_{16}$ & \modelname{} (Top1/1)
    & 5.57M & 0.043
    & 64.98 & 78.56 & 72.52 & 41.99 & 67.25 & 77.82 & 58.70 & 68.20 & 66.25 \\
    LoRA$_{32}$ & \modelname{} (Top1/1)
    & 1.08M & 0.084
    & 66.36 & 78.84 & 72.36 & 42.83 & 63.38 & 78.62 & 58.36 & 71.20 & 66.49 \\
\midrule
    LoRA$_{4}$ & \modelname{} (Top1/2)
    & 0.20M & 0.015
    & 63.67 & 77.04 & 69.09 & 39.92 & 58.09 & 76.81 & 55.80 & 62.40 & 62.85 \\
    LoRA$_{8}$ & \modelname{} (Top1/2)
    & 0.33M & 0.026
    & 63.98 & 78.13 & 70.93 & 41.00 & 58.88 & 78.11 & 56.66 & 65.80 & 64.19 \\
    LoRA$_{16}$ & \modelname{} (Top1/2)
    & 0.59M & 0.046
    & 65.14 & 76.93 & 72.42 & 41.39 & 70.64 & 78.03 & 59.56 & 69.20 & 66.66 \\
    LoRA$_{32}$ & \modelname{} (Top1/2)
    & 1.11M & 0.087
    & 65.60 & 78.18 & 73.13 & 43.47 & 69.61 & 77.40 & 58.53 & 70.00 & 66.99 \\
    LoRA$_{64}$ & \modelname{} (Top1/2)
    & 2.16M & 0.169
    & 66.09 & 77.97 & 73.75 & 46.36 & 72.61 & 78.79 & 62.20 & 69.20 & 68.37 \\
\midrule
    LoRA$_{4}$ & \modelname{} (Top2/2)
    & 0.33M & 0.026
    & 64.86 & 76.71 & 69.60 & 40.89 & 62.43 & 77.23 & 55.80 & 63.60 & 63.89 \\
    LoRA$_{8}$ & \modelname{} (Top2/2)
    & 0.59M & 0.046
    & 65.26 & 78.18 & 72.31 & 42.11 & 71.82 & 77.90 & 60.49 & 67.80 & 66.99 \\
    LoRA$_{16}$ & \modelname{} (Top2/2)
    & 1.11M & 0.087
    & 66.18 & 77.97 & 72.52 & 43.99 & 70.64 & 78.24 & 60.75 & 69.80 & 67.51 \\
    LoRA$_{32}$ & \modelname{} (Top2/2)
    & 2.16M & 0.169
    & 65.81 & 79.38 & 73.59 & 49.42 & 71.59 & 77.78 & 61.18 & 71.80 & 68.82 \\
    LoRA$_{64}$ & \modelname{} (Top2/2)
    & 4.26M & 0.332
    & 65.96 & 79.87 & 72.82 & 53.93 & 73.40 & 78.91 & 62.20 & 72.20 & 69.91 \\
    LoRA$_{128}$ & \modelname{} (Top2/2)
    & 8.45M & 0.659
    & 67.09 & 80.09 & 74.67 & 68.44 & 70.32 & 79.55 & 60.49 & 73.80 & 71.81 \\
\midrule    
    LoRA$_{4}$ & \modelname{} (Top1/4)
    & 0.39M & 0.031
    & 63.94 & 76.88 & 69.91 & 39.14 & 60.54 & 78.49 & 57.68 & 65.40 & 64.00 \\
    LoRA$_{8}$ & \modelname{} (Top1/4)
    & 0.66M & 0.051
    & 64.34 & 77.75 & 71.75 & 40.30 & 67.01 & 77.06 & 58.96 & 64.80 & 65.25 \\
    LoRA$_{16}$ & \modelname{} (Top1/4)
    & 1.18M & 0.092
    & 64.46 & 77.04 & 71.29 & 41.83 & 62.51 & 77.57 & 59.39 & 65.00 & 64.89 \\
    LoRA$_{32}$ & \modelname{} (Top1/4)
    & 2.23M & 0.174
    & 66.21 & 78.51 & 71.49 & 43.87 & 69.61 & 77.69 & 61.01 & 70.20 & 67.32 \\
    LoRA$_{64}$ & \modelname{} (Top1/4)
    & 4.33 & 0.337
    & 65.32 & 79.60 & 73.49 & 45.33 & 71.11 & 77.69 & 62.20 & 71.00 & 68.22 \\
\midrule
    LoRA$_{4}$ & \modelname{} (Top2/4)
    & 0.66M & 0.051
    & 63.98 & 75.68 & 69.29 & 40.26 & 65.75 & 77.36 & 59.56 & 67.40 & 64.91 \\
    LoRA$_{8}$ & \modelname{} (Top2/4)
    & 1.18M & 0.092
    & 65.02 & 77.86 & 71.90 & 41.61 & 68.75 & 77.31 & 59.13 & 68.80 & 66.30 \\
    LoRA$_{16}$ & \modelname{} (Top2/4)
    & 2.23M & 0.174
    & 64.07 & 76.61 & 73.59 & 42.10 & 71.90 & 78.32 & 60.58 & 71.20 & 67.30 \\
    LoRA$_{32}$ & \modelname{} (Top2/4)
    & 4.33M & 0.337
    & 66.30 & 77.75 & 75.44 & 45.88 & 71.43 & 76.18 & 60.58 & 70.60 & 68.02 \\
\midrule
    LoRA$_{4}$ & \modelname{} (Top4/4)
    & 1.18M & 0.092
    & 64.25 & 75.84 & 71.03 & 41.40 & 69.22 & 77.65 & 57.08 & 68.40 & 65.61 \\
    LoRA$_{8}$ & \modelname{} (Top4/4)
    & 2.23M & 0.174
    & 65.14 & 77.64 & 72.98 & 42.67 & 72.45 & 76.98 & 59.39 & 66.40 & 66.71 \\
    LoRA$_{16}$ & \modelname{} (Top4/4)
    & 4.33M & 0.337
    & 65.44 & 79.43 & 73.08 & 48.35 & 71.19 & 77.48 & 59.98 & 73.40 & 68.55 \\
    LoRA$_{32}$ & \modelname{} (Top4/4)
    & 8.52M & 0.665
    & 66.70 & 79.49 & 73.75 & 55.95 & 71.43 & 77.53 & 60.07 & 70.40 & 69.41 \\
    LoRA$_{64}$ & \modelname{} (Top4/4)
    & 16.9M & 1.319
    & 66.02 & 79.71 & 75.49 & 59.29 & 73.32 & 76.64 & 59.90 & 71.80 & 70.27 \\
    LoRA$_{128}$ & \modelname{} (Top4/4)
    & 33.7M & 2.628
    & 65.99 & 78.94 & 75.13 & 67.21 & 73.72 & 78.24 & 59.90 & 74.80 & 71.74 \\
\midrule
\caption{\textbf{(Part 1/2) Evaluation results for OLMoE with baseline methods and \modelname{} variants on eight commonsense reasoning benchmarks.} ``Arch.'' denotes the architecture inside PEFT modules. ``\# Act.'' and ``\% Act.'' represent the number of activated trainable parameters and their ratio to the total activated parameters. ``(TopK/N)'' refers to activating $K$ experts among the total number of $N$ experts. Dataset names are partially abbreviated, including BoolQ \citep{clark2019boolq}, PIQA \citep{bisk2020piqa}, Social IQa \citep{sap2019siqa}, HellaSwag \citep{zellers2019hellaswag}, WinoGrande \citep{sakaguchi2021winogrande}, Easy Set and Challenge Set of ARC \citep{clark2018arc}, and OpenBookQA \citep{mihaylov2018obqa}.
}
\label{tab:result-commonsense-olmoe-1}
\end{longtable}
}

\newpage

{\renewcommand{\arraystretch}{0.9}
\centering
\scriptsize
\addtolength{\tabcolsep}{-3pt} 
\begin{longtable}{ll|cc|cccccccc|c}
\toprule
    \textbf{Arch.} & \textbf{Strategy }
    & \textbf{\# Act.} & \textbf{\% Act.}
    & \textbf{BoolQ} & \textbf{PIQA}  & \textbf{SIQA}  & \textbf{HellaS}
    & \textbf{WinoG} & \textbf{ARC-e} & \textbf{ARC-c} & \textbf{OBQA}  & \textbf{Avg.}\\  
\midrule
    LoRA$_{4}$ & \modelname{} (Top1/8)
    & 0.52M & 0.041
    & 63.73 & 75.30 & 69.91 & 40.77 & 66.77 & 77.69 & 57.51 & 64.60 & 64.54 \\
    LoRA$_{8}$ & \modelname{} (Top1/8)
    & 0.79M & 0.061
    & 64.98 & 77.09 & 70.78 & 41.65 & 66.93 & 77.78 & 57.76 & 66.40 & 65.42 \\
    LoRA$_{16}$ & \modelname{} (Top1/8)
    & 1.31M & 0.102
    & 64.89 & 77.26 & 70.88 & 41.95 & 70.09 & 77.31 & 59.39 & 67.40 & 66.15 \\
    LoRA$_{32}$ & \modelname{} (Top1/8)
    & 2.36M & 0.184
    & 64.25 & 77.58 & 72.52 & 42.30 & 70.64 & 77.82 & 58.53 & 67.40 & 66.38 \\
\midrule
    LoRA$_{4}$ & \modelname{} (Top2/8)
    & 0.79M & 0.061
    & 64.28 & 76.99 & 68.88 & 40.61 & 66.85 & 77.57 & 57.34 & 65.40 & 64.74 \\
    LoRA$_{8}$ & \modelname{} (Top2/8)
    & 1.31M & 0.102
    & 63.91 & 76.88 & 71.03 & 43.45 & 69.69 & 77.23 & 58.11 & 68.00 & 66.04 \\
    LoRA$_{16}$ & \modelname{} (Top2/8)
    & 2.36M & 0.184
    & 64.68 & 77.64 & 72.36 & 43.33 & 71.51 & 75.97 & 58.45 & 67.80 & 66.47 \\
    LoRA$_{32}$ & \modelname{} (Top2/8)
    & 4.46M & 0.348
    & 64.40 & 78.13 & 74.21 & 46.80 & 71.59 & 76.39 & 58.79 & 71.20 & 67.69 \\
\midrule
    LoRA$_{4}$ & \modelname{} (Top4/8)
    & 1.31M & 0.102
    & 64.74 & 77.04 & 71.60 & 42.82 & 70.01 & 77.31 & 59.73 & 68.20 & 66.43 \\
    LoRA$_{8}$ & \modelname{} (Top4/8)
    & 2.36M & 0.184
    & 64.86 & 76.61 & 73.69 & 42.10 & 69.46 & 76.98 & 58.02 & 67.20 & 66.12 \\
    LoRA$_{16}$ & \modelname{} (Top4/8)
    & 4.46M & 0.348
    & 65.78 & 76.33 & 72.57 & 45.61 & 69.53 & 76.22 & 58.28 & 69.20 & 66.69 \\
    LoRA$_{32}$ & \modelname{} (Top4/8)
    & 8.65M & 0.675
    & 65.20 & 77.37 & 73.64 & 46.36 & 72.45 & 77.02 & 56.83 & 69.20 & 67.26 \\
\midrule
    LoRA$_{4}$ & \modelname{} (Top8/8)
    & 2.36M & 0.184 
    & 64.98 & 77.37 & 72.77 & 45.71 & 70.32 & 77.15 & 58.96 & 68.60 & 66.98 \\
    LoRA$_{8}$ & \modelname{} (Top8/8)
    & 4.46M &  0.348
    & 64.98 & 78.13 & 74.21 & 46.75 & 69.85 & 77.19 & 59.56 & 70.00 & 67.58 \\
    LoRA$_{16}$ & \modelname{} (Top8/8)
    & 8.65M & 0.675 
    & 65.93 & 77.58 & 74.41 & 55.14 & 71.98 & 76.47 & 57.59 & 71.40 & 68.81 \\
    LoRA$_{32}$ & \modelname{} (Top8/8)
    & 17.0M &  1.329
    & 65.78 & 78.07 & 74.92 & 58.44 & 71.82 & 76.05 & 61.35 & 73.80 & 70.03 \\
    LoRA$_{64}$ & \modelname{} (Top8/8)
    & 33.8M & 2.638 
    & 65.20 & 80.25 & 75.13 & 65.68 & 73.01 & 75.67 & 59.47 & 72.40 & 70.85 \\
\midrule
    LoRA$_{4}$ & \modelname{} (Top1/16)
    & 0.79M & 0.061
    & 64.65 & 75.73 & 70.83 & 40.04 & 63.61 & 77.06 & 59.04 & 64.40 & 64.42 \\
    LoRA$_{8}$ & \modelname{} (Top1/16)
    & 1.05M & 0.082
    & 64.98 & 76.17 & 69.60 & 40.17 & 67.48 & 76.30 & 58.02 & 67.00 & 64.97 \\
    LoRA$_{16}$ & \modelname{} (Top1/16)
    & 1.57M & 0.123
    & 63.79 & 77.04 & 73.29 & 42.39 & 70.56 & 76.60 & 58.96 & 69.00 & 66.45 \\
    LoRA$_{32}$ & \modelname{} (Top1/16)
    & 2.62M & 0.204
    & 64.25 & 75.79 & 72.21 & 43.98 & 70.24 & 76.18 & 59.04 & 69.20 & 66.36 \\
\midrule
    LoRA$_{4}$ & \modelname{} (Top2/16)
    & 1.05M & 0.082
    & 63.94 & 77.31 & 71.44 & 41.23 & 69.22 & 78.37 & 58.11 & 67.00 & 65.83 \\
    LoRA$_{8}$ & \modelname{} (Top2/16)
    & 1.57M & 0.123
    & 62.45 & 76.12 & 71.55 & 41.75 & 67.80 & 76.14 & 59.47 & 68.00 & 65.41 \\
    LoRA$_{16}$ & \modelname{} (Top2/16)
    & 2.62M & 0.204
    & 64.50 & 76.06 & 71.03 & 43.21 & 69.22 & 75.59 & 59.30 & 68.00 & 65.86 \\
    LoRA$_{32}$ & \modelname{} (Top2/16)
    & 4.72M & 0.368
    & 65.35 & 76.50 & 72.98 & 47.08 & 69.30 & 74.79 & 58.19 & 67.80 & 66.50 \\
\midrule
    LoRA$_{4}$ & \modelname{} (Top4/16)
    & 1.57M & 0.123
    & 64.37 & 75.52 & 72.36 & 42.12 & 69.61 & 76.35 & 57.59 & 68.00 & 65.74 \\
    LoRA$_{8}$ & \modelname{} (Top4/16)
    & 2.62M & 0.204
    & 64.92 & 76.55 & 72.21 & 43.09 & 69.61 & 75.67 & 59.30 & 67.20 & 66.07 \\
    LoRA$_{16}$ & \modelname{} (Top4/16)
    & 4.72M & 0.368
    & 65.50 & 76.50 & 73.80 & 43.82 & 71.43 & 74.03 & 57.34 & 69.80 & 66.53 \\
    LoRA$_{32}$ & \modelname{} (Top4/16)
    & 8.91M & 0.695
    & 65.47 & 77.09 & 73.64 & 45.04 & 69.77 & 74.49 & 58.70 & 67.80 & 66.50 \\
\midrule
    LoRA$_{4}$ & \modelname{} (Top8/16)
    & 2.62M & 0.204
    & 64.25 & 76.06 & 72.31 & 41.46 & 71.11 & 76.81 & 60.67 & 68.00 & 66.33 \\
    LoRA$_{8}$ & \modelname{} (Top8/16)
    & 4.72M & 0.368
    & 64.50 & 77.53 & 73.34 & 45.22 & 71.74 & 74.92 & 57.51 & 67.80 & 66.57 \\
    LoRA$_{16}$ & \modelname{} (Top8/16)
    & 8.91M & 0.695
    & 64.53 & 77.91 & 73.54 & 47.24 & 71.27 & 75.00 & 54.78 & 71.20 & 66.93 \\
    LoRA$_{32}$ & \modelname{} (Top8/16)
    & 17.3M & 1.350
    & 65.57 & 76.82 & 74.51 & 53.13 & 70.01 & 74.07 & 57.17 & 70.60 & 67.73 \\
\midrule
    LoRA$_{4}$ & \modelname{} (Top8/32)
    & 3.15M & 0.245
    & 63.82 & 75.52 & 72.57 & 41.75 & 72.30 & 74.37 & 57.25 & 69.00 & 65.82 \\
    LoRA$_{8}$ & \modelname{} (Top8/32)
    & 5.24M & 0.409
    & 63.79 & 75.35 & 71.70 & 43.90 & 67.88 & 74.03 & 58.28 & 67.80 & 65.34 \\
    LoRA$_{16}$ & \modelname{} (Top8/32)
    & 9.44M & 0.736
    & 64.07 & 75.90 & 73.39 & 44.59 & 72.22 & 72.31 & 55.29 & 65.20 & 65.37 \\
    LoRA$_{32}$ & \modelname{} (Top8/32)
    & 17.8M & 1.390
    & 64.71 & 75.35 & 73.95 & 47.17 & 70.72 & 72.22 & 55.46 & 67.80 & 65.92 \\
\midrule
    LoRA$_{4}$ & \modelname{} (Top8/64)
    & 4.19M & 0.327
    & 63.55 & 76.06 & 70.11 & 42.16 & 69.14 & 72.31 & 53.67 & 64.80 & 63.98 \\
    LoRA$_{8}$ & \modelname{} (Top8/64)
    & 6.29M & 0.491
    & 64.53 & 75.52 & 72.21 & 41.79 & 70.40 & 71.38 & 53.92 & 66.20 & 64.49 \\
    LoRA$_{16}$ & \modelname{} (Top8/64)
    & 10.5M & 0.818
    & 64.71 & 73.61 & 72.26 & 42.35 & 70.88 & 71.09 & 54.78 & 65.80 & 64.44 \\
    LoRA$_{32}$ & \modelname{} (Top8/64)
    & 18.9M & 1.472
    & 62.81 & 74.43 & 72.31 & 41.11 & 69.22 & 69.49 & 53.84 & 65.60 & 63.60 \\
\midrule
\midrule
    LoRA$_{2}$ & \modelname{}-E (Top8/64)
    & 1.05M & 0.082
    & 65.54 & 79.11 & 73.59 & 50.06 & 73.24 & 77.27 & 58.70 & 72.80 & 68.79 \\
    LoRA$_{4}$ & \modelname{}-E (Top8/64)
    & 2.10M & 0.164
    & 64.80 & 79.49 & 74.36 & 58.39 & 72.69 & 75.00 & 58.45 & 72.20 & 69.42 \\
    LoRA$_{8}$ & \modelname{}-E (Top8/64)
    & 4.19M & 0.327
    & 65.81 & 78.84 & 73.85 & 58.84 & 71.51 & 74.41 & 56.06 & 69.20 & 68.56 \\
    LoRA$_{16}$ & \modelname{}-E (Top8/64)
    & 8.39M & 0.654
    & 65.20 & 78.24 & 74.97 & 64.35 & 72.30 & 74.41 & 55.46 & 69.40 & 69.29 \\
    LoRA$_{32}$ & \modelname{}-E (Top8/64)
    & 16.8M & 1.309
    & 66.51 & 76.39 & 74.26 & 62.55 & 73.09 & 72.22 & 56.14 & 70.60 & 68.97 \\
    LoRA$_{64}$ & \modelname{}-E (Top8/64)
    & 33.6M & 2.617
    & 65.57 & 77.09 & 73.80 & 59.89 & 73.32 & 71.72 & 56.40 & 68.80 & 68.32 \\
\bottomrule
\caption{\textbf{(Part 2/2) Evaluation results for \olmoe{} with baseline methods and \modelname{} variants on eight commonsense reasoning benchmarks.} ``Arch.'' denotes the architecture inside PEFT modules. ``\# Act.'' and ``\% Act.'' represent the number of activated trainable parameters and their ratio to the total activated parameters. ``(TopK/N)'' refers to activating $K$ experts among the total number of $N$ experts. Dataset names are partially abbreviated, including BoolQ \citep{clark2019boolq}, PIQA \citep{bisk2020piqa}, Social IQa \citep{sap2019siqa}, HellaSwag \citep{zellers2019hellaswag}, WinoGrande \citep{sakaguchi2021winogrande}, Easy Set and Challenge Set of ARC \citep{clark2018arc}, and OpenBookQA \citep{mihaylov2018obqa}.
}
\label{tab:result-commonsense-olmoe-2}
\end{longtable}
}

\newpage

\subsection{\mixtral{} for Commonsense Reasoning}

{\renewcommand{\arraystretch}{0.9}
\centering
\scriptsize
\addtolength{\tabcolsep}{-3pt} 
\begin{longtable}{ll|cc|cccccccc|c}
\toprule
    \textbf{Arch.} & \textbf{Strategy }
    & \textbf{\# Act.} & \textbf{\% Act.}
    & \textbf{BoolQ} & \textbf{PIQA}  & \textbf{SIQA}  & \textbf{HellaS}
    & \textbf{WinoG} & \textbf{ARC-e} & \textbf{ARC-c} & \textbf{OBQA}  & \textbf{Avg.}\\  
\midrule
    Base & (pretrained)
    & --- & ---
    & 51.10 & 81.12 & 46.11 & 47.54 & 49.88 & 53.20 & 52.99 & 39.20 & 52.64 \\
    Base & (instruct)
    & --- & ---
    & 68.87 & 88.30 & 68.58 & 72.06 & 59.98 & 89.52 & 78.50 & 74.40 & 75.03\\
\midrule
\midrule
    LoRA$_{8}$  & $\bm{W}_q, \bm{W}_v$@$\texttt{Attn}$
    & 3.41M  & 0.026
    & 73.49 & 90.04 & 81.17 & 89.67 & 82.16 & 93.56 & 83.87 & 86.20 & 85.02 \\
\midrule
\midrule
    LoRA$_{16}$ & \modelname{}-S (1)
    & 4.19M & 0.033
    & 75.11 & 90.26 & 81.63 & 94.26 & 84.85 & 92.85 & 81.40 & 87.60 & 85.99 \\
\midrule
\midrule
    LoRA$_{8}$ & \modelname{} (Top2/2)
    & 4.46M  & 0.035
    & 74.68 & 89.77 & 81.47 & 94.33 & 86.27 & 92.05 & 81.48 & 89.80 & 86.23 \\
    LoRA$_{16}$ & \modelname{} (Top1/4)
    & 4.72M & 0.037
    & 72.84 & 89.12 & 80.40 & 92.69 & 84.37 & 91.84 & 82.25 & 85.80 & 84.91 \\
    LoRA$_{8}$ & \modelname{} (Top2/4)
    & 4.72M & 0.037
    & 74.71 & 90.10 & 79.38 & 94.18 & 85.71 & 92.09 & 81.31 & 85.80 & 85.41 \\
    LoRA$_{8}$ & \modelname{} (Top2/8)
    & 5.24M  & 0.041
    & 73.76 & 89.12 & 81.63 & 94.51 & 85.16 & 91.67 & 80.20 & 87.80 & 85.48 \\
\midrule
\midrule
    LoRA$_{8}$ & \modelname{}-E (Top2/8)
    & 4.19M  & 0.033
    & 74.13 & 90.21 & 80.81 & 91.36 & 86.42 & 92.21 & 81.06 & 88.60 & 85.60 \\
\bottomrule
\caption{\textbf{Evaluation results for \mixtral{} with baseline methods and \modelname{} variants on eight commonsense reasoning benchmarks.} ``Arch.'' denotes the architecture inside PEFT modules. ``\# Act.'' and ``\% Act.'' represent the number of activated trainable parameters and their ratio to the total activated parameters. ``(TopK/N)'' refers to activating $K$ experts among the total number of $N$ experts. Dataset names are partially abbreviated, including BoolQ \citep{clark2019boolq}, PIQA \citep{bisk2020piqa}, Social IQa \citep{sap2019siqa}, HellaSwag \citep{zellers2019hellaswag}, WinoGrande \citep{sakaguchi2021winogrande}, Easy Set and Challenge Set of ARC \citep{clark2018arc}, and OpenBookQA \citep{mihaylov2018obqa}.
}
\label{tab:result-commonsense-mixtral}
\end{longtable}
}

\subsection{\mixtral{} for Arithmetic Reasoning}

{\renewcommand{\arraystretch}{0.9}
\centering
\scriptsize
\addtolength{\tabcolsep}{-2.5pt} 
\begin{longtable}{ll|cc|cccccc|c}
\toprule
    \textbf{Arch.} & \textbf{Strategy }
    & \textbf{\# Act.} & \textbf{\% Act.}
    & \textbf{MultiArith} & \textbf{GSM8K}  & \textbf{AddSub}  & \textbf{AQuA}
    & \textbf{SingleEq} & \textbf{SVAMP} & \textbf{Avg.}\\  
\midrule
    LoRA$_{8}$  & $\bm{W}_q, \bm{W}_v$@$\texttt{Attn}$
    & 3.41M  & 0.026
    & 60.00 & 50.87 & 90.13 & 28.74 & 89.37 & 69.20 & 64.72 \\
\midrule
\midrule
    LoRA$_{8}$ & \modelname{} (Top2/2)
    & 4.46M  & 0.035
    & 82.83 & 55.80 & 87.59 & 29.92 & 89.76 & 68.30 & 69.04 \\
    LoRA$_{8}$ & \modelname{} (Top2/8)
    & 5.24M  & 0.041
    & 79.00 & 54.06 & 87.34 & 29.13 & 88.98 & 70.30 & 68.13 \\
\bottomrule
\caption{\textbf{Evaluation results for \mixtral{} with baseline methods and \modelname{} variants on six arithmetic reasoning benchmarks.} ``Arch.'' denotes the architecture inside PEFT modules. ``\# Act.'' and ``\% Act.'' represent the number of activated trainable parameters and their ratio to the total activated parameters. ``(TopK/N)'' refers to activating $K$ experts among the total number of $N$ experts. Dataset names are partially abbreviated, including MultiArith \citep{roy2015solving}, GSM8K \citep{cobbe2021training}, AddSub \citep{hosseini2014learning}, AQuA \citep{ling2017program}, SingleEq \citep{koncel2015parsing}, and SVAMP \citep{patel2021nlp}.
}
\label{tab:result-math-mixtral}
\end{longtable}
}

\end{document}